\documentclass{article} 
\usepackage{iclr2025_conference,times}

\usepackage{amsmath,amsfonts,bm}

\def\eqref#1{equation~\ref{#1}}

\def\1{\bm{1}}

\DeclareMathAlphabet{\mathsfit}{\encodingdefault}{\sfdefault}{m}{sl}
\SetMathAlphabet{\mathsfit}{bold}{\encodingdefault}{\sfdefault}{bx}{n}

\usepackage{hyperref}
\usepackage{url}
\usepackage{graphicx}
\newcommand{\our}{DiffCoRe-Mix}
\usepackage{ulem}
\usepackage{multirow}

\usepackage{xcolor}
\usepackage{graphicx}
\usepackage{amsmath}
\usepackage{booktabs}
\usepackage{pifont}    

\usepackage{color}
\usepackage{colortbl}
\usepackage{array}
\usepackage{soul}
\usepackage{balance}
 \usepackage{amssymb} 
\usepackage{wrapfig}
\usepackage{subcaption}

\newcommand{\numbersBlue}[1]{\textcolor{blue}{\textbf{#1}}}
\definecolor{mygray}{rgb}{.95,.95,.95}
\definecolor{myblue}{rgb}{0.85, 0.9, 1.0}
\definecolor{suppColor}{RGB}{148, 0, 211} 
\usepackage{colortbl} 
\usepackage{booktabs} 
\usepackage{xcolor}   

\newcommand{\yes}{\textcolor{darkgray}{\ding{51}}}

\usepackage{caption}
\definecolor{darkgreen}{rgb}{0.0, 0.5, 0.0}

\title{Context-guided Responsible Data Augmentation with Diffusion Models}

\iclrfinalcopy 

\author{Khawar Islam, Naveed Akhtar \\
School of Computing and Information Systems, The University of Melbourne \\
Melbourne, Parkville VIC 3010, Australia\\
\texttt{khawar.islam@student.unimelb.edu.au,  naveed.akhtar1@unimelb.edu.au} \\
}

\begin{document}

\maketitle

\begin{abstract}

Generative diffusion models offer a natural choice for data augmentation when training complex vision models.  
However, ensuring reliability of their generative content as augmentation samples remains an open challenge. Despite a number of techniques utilizing generative images to strengthen model  training,  it remains unclear how to utilize the combination of natural and generative images as a rich supervisory signal for effective model induction. In this regard, we propose a text-to-image (T2I) data augmentation method, named \our{}, that computes a set of generative counterparts for a training sample with an explicitly constrained diffusion model that leverages  sample-based  context and negative prompting for a reliable augmentation sample generation. To preserve key semantic axes, we also filter out undesired generative samples in our augmentation process. To that end, we propose a hard-cosine filtration in the embedding space of CLIP.  Our approach systematically mixes the natural and generative images at pixel  and patch levels. We extensively evaluate our technique on \textit{ImageNet-1K}, \textit{Tiny ImageNet-200}, \textit{CIFAR-100},  \textit{Flowers102}, \textit{CUB-Birds}, \textit{ Stanford Cars}, and \textit{Caltech} datasets, demonstrating a notable increase in performance  across the board, achieving up to $\sim 3\%$ absolute gain for top-1 accuracy  over the state-of-the-art methods, while showing comparable computational overhead. Our code is publicly available at \href{https://github.com/khawar-islam/DiffCoRe-Mix}{\numbersBlue{DiffCoRe-Mix}}.

\end{abstract}

\section{Introduction}
\label{sec:intro}
\vspace{-1.5mm}
Mixup data augmentation methods ~\citep{kim2020co, kang2023guidedmixup} are widely used to augment training data of neural models to achieve better generalization. Approaches under this paradigm  devise sophisticated mechanisms to mix  different images using apriori or saliency information \citep{qin2024sumix, qin2023adversarial, han2022yoco, chen2022transmix, choi2022tokenmixup}. Though effective, these techniques must overcome a critical inherent limitation of the paradigm, which requires deciding on an appropriate supervisory signal for the added augmentation samples~\citep{islam2024diffusemix}. Ambiguity in this signal can even lead to reducing model generalization instead of improving it~\citep{azizi2023synthetic}.

Recently, Diffusion Models (DMs) have shown remarkable abilities of generating high quality realistic images~\citep{rombach2022high, richhf, meral2024conform}. 
Conditioned on an image or text, DMs can generate multiple new images for a given class by using class-label information in their prompts. Using such images as added training data has emerged as an effective alternate to the conventional data augmentation strategy of using input transformations as the added samples \citep{islam2024diffusemix, trabucco2023effective, tian2024stablerep, fu2024dreamda, luo2023camdiff}. Nevertheless, this alternative comes with its own challenges - the central problem being the inadequate control over the content of the generated images, which can lead to ineffective or even detrimental samples. 
\par
Currently, gaining better control over the generative content in DMs is emerging as an active parallel  research direction ~\citep{mou2024diffeditor, shi2024dragdiffusion, huang2021snapmix, xu2024inversion}. However, it is yet to focus on achieving semantic coherence and appropriate alignment with the original data samples for the purpose of data augmentation, which is still widely open. The early DM based augmentation methods  \citep{hesynthetic, trabucco2023effective, wang2024enhance} mainly trusted the impromptu generative outputs for augmentation.  Addressing this inadequacy, there are recent attempts to use image editing with DMs for augmentation \citep{brooks2023instructpix2pix, tian2024stablerep}. 
However, these techniques largely overlook the advances in the traditional image-mixing paradigm, thereby falling short on fully exploiting them. An exception to that is \citep{islam2024diffusemix, islam2024genmix}, which proposes to leverage traditional image-mixing with an image-to-image (I2I) generative model for data augmentation.  

\begin{table*}[t]
\centering
\renewcommand{\arraystretch}{1.0}
\caption{Comparison of representative mixup, generative methods and generative mixup data augmentation methods. }

\scalebox{0.78}{
\begin{tabular}{lccccccc}
\toprule
\rowcolor{mygray}
\textbf{} & \multicolumn{2}{c}{Mixup Methods} & \multicolumn{3}{c}{Generative Methods} & \multicolumn{2}{c}{Generative Mixup Methods} 
\\ 
\cmidrule(lr){2-3}  \cmidrule(lr){4-6} \cmidrule(lr){7-8}
\rowcolor{mygray} & CoMixup & Guided-AP  & Real-Guid & DA-Fusion   & Diff-Mix & DiffuseMix & \textbf{\our{}} 
\\ 
\midrule
\textbf{Mixing} & Saliency & Saliency & \textemdash & \textemdash & \textemdash & Mask-Wise & P- \& P-Wise 
\\ \arrayrulecolor{lightgray}\midrule
\multirow{2}{*}{\textbf{Prompt (P)}} & \textemdash & \textemdash & \multirow{2}{*}{\begin{tabular}[c]{@{}c@{}}Label \\ Description \end{tabular}} & \multirow{2}{*}{\begin{tabular}[c]{@{}c@{}}Derived from \\ Intra-Class \end{tabular}} & \multirow{2}{*}{\begin{tabular}[c]{@{}c@{}}Derived from \\ Inter-Class \end{tabular}} & Style Prompt & \multirow{2}{*}{\begin{tabular}[c]{@{}c@{}} General\end{tabular}} \\
& \multicolumn{2}{c}{} & & & & & 
\\  \arrayrulecolor{lightgray}\midrule

\textbf{Negative P} & \textemdash & \textemdash & \textemdash & \textemdash & \textemdash & \textemdash & \yes 
\\ \arrayrulecolor{lightgray}\midrule

\textbf{Contextual P} & \textemdash & \textemdash & \textemdash & \textemdash & \textemdash & \textemdash & \yes \\ 




\arrayrulecolor{black}\bottomrule
\end{tabular}
}
\label{table:compar_study}
\end{table*}

\begin{wrapfigure}{r}{0.5\textwidth}
    \centering
    \includegraphics[width=0.50\textwidth]{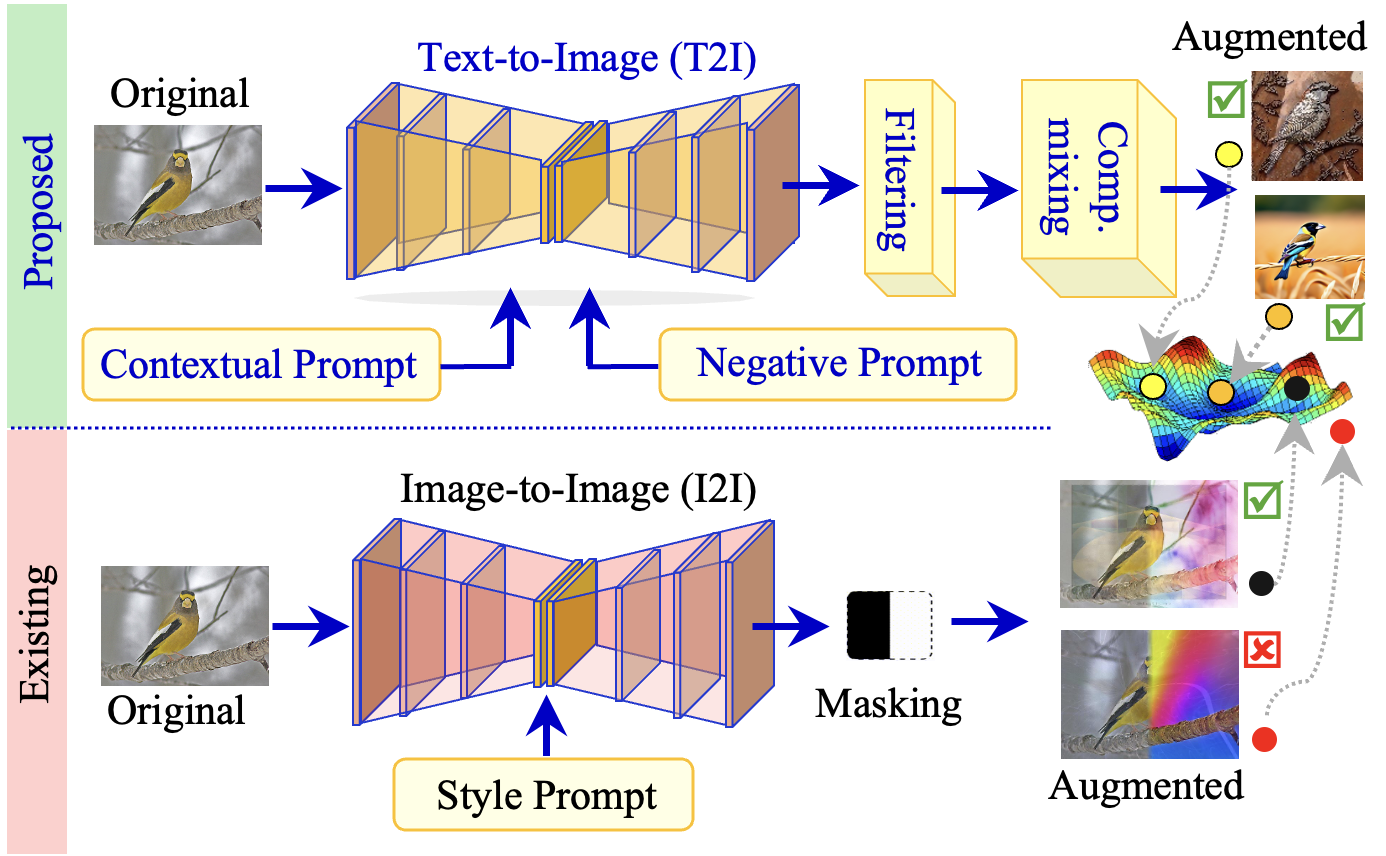}
    \captionsetup{width=0.50\textwidth} 
    \caption{(\textbf{Top}) The proposed \our{} employs a T2I model constrained with contextual and negative prompts. The output of the T2I model is filtered, and image-mixing is employed to introduce better generalization and robustness.(\textbf{Bottom}) The closest generative image-mixing method~\citep{islam2024diffusemix} uses an I2I model with style prompt to edit the image by concatenating original and generative image.}
    \label{fig:comparison}
\end{wrapfigure}

With our text prompts, we ensure an improved  control over the outputs by contextualizing the prompt with the original image label. As an additional supervisory signal, we also employ Negative Prompts to restrict the generative output space of the diffusion model. Moreover, we additionally filter out any undesirable generative outputs in a fully automated manner. To that end, we develop a hard cosine filtration mechanism that is deployed in the embedding space of the CLIP-encoder. This filtration affirms appropriateness of the generative image set. We employee patch-level regularization and pixel-level sensitivity based mixing of the original and the generative images to construct the augmented data for improved model performance. Our \ul{Co}ntext-guided \ul{Diff}usion based  method enables \ul{Re}sponsible image-\ul{Mix}ing in  that the augmentation samples align well  with the original data  - hence  termed DiffCoRe-Mix.  Over the closest technique~\citep{islam2024diffusemix}, it provides a strong advantage of avoiding unrealistic or ill-formed augmentation samples - see Fig.~\ref{fig:comparison}, which results from foundational technical differences. Over other  data augmentation methods, it provides different  advantages - see Tab.~\ref{table:compar_study}, along with stronger   performance. Our main contributions are summarized below.

\begin{itemize}
    \item We propose T2I generative data augmentation that ensures semantic alignment of the generative image with the original image while preserving fine-grained details. 
    
    \item We introduce contextual and negative prompting to ensure domain-specific generative images while restricting undesired samples, and also devise a hard cosine similarity filtration for the CLIP embedding space to further semantically align the generative images to the original samples. 
    
    \item We incorporate real and generative image into pixel wise approach to reduce the memorization of neural network, and patch-wise to enhance regularization.
    
    \item We establish notable efficacy of our approach  with extensive experiments on six datasets for the tasks including general classification, fine-grained classification, fine-tuning, and data scarcity; outperforming the state-of-the-art methods across the board by a considerable margin.
\end{itemize}

\begin{figure*}[t]
    \centering
    \includegraphics[width=1.0\linewidth]{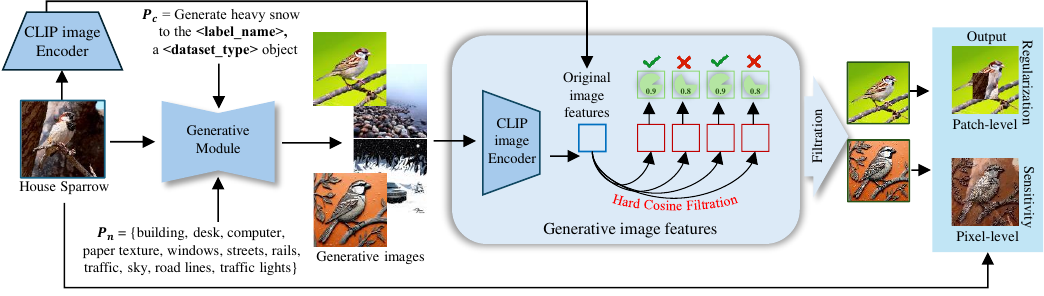}
    \caption{Overview of \our{} data augmentation method. It takes an input image from dataset to generate a image guided by our contextual and negative prompts. CLIP-based image encoder is utilized to extract features from original and generative image. Then, our hard-cosine filtration approach is used to verify the semantic alignment between the original and generative features. We filter out unaligned images, and mix pixel- and patch-level real and generative images.}
    \label{fig:overview}
\end{figure*}

\section{Related Work}
Below,  recent advances in data augmentation for vision models are discussed while focusing on the key methods related to our approach.
\par
\noindent{\textbf{Mixup Methods:}} 
It is widely known that data augmentation helps in model generalization. Deep learning methods commonly apply basic transformations to inputs, e.g., rotation, flipping,  to construct augmentation samples. However, more recently, dedicated approaches have emerged for a sophisticated data augmentation  \citep{kang2023guidedmixup, qin2024sumix, qin2023adversarial, han2022yoco, chen2022transmix, choi2022tokenmixup}. Among them, image-mixing is the paradigm that mixes a given input sample with other training  samples or their sub-parts to create augmentation samples. For instance, Mixup  \citep{mixup} mixes two random images to reduce memorization and increase generalization of classifiers. Similarly, CutMix \citep{cutmix} randomly cuts and pastes portions of images to improve model performance on out-of-distribution samples. ResizeMix \citep{qin2020resizemix} modifies CutMix \citep{cutmix} by resizing image sections instead of cutting them, offering a smoother blending. SmoothMix \citep{lee2020smoothmix} makes patch  boundaries smoother for better blending to improve mixing.  In another line of work, methods like SaliencyMix \citep{uddin2020saliencymix} and Attentive-CutMix \citep{walawalkar2020attentive} use saliency extraction to blend the most crucial parts of the images. PuzzleMix \citep{kim2020puzzle}, GuidedMixup \citep{kang2023guidedmixup} and Co-Mixup \citep{kim2020co} take this a step further by isolating important regions from both source and target images. Co-Mixup \citep{kim2020co} introduces more sophistication by mixing three images instead of two. AutoMix \citep{liu2022automix} and SAMix \citep{li2021boosting} explore the balance between hand-crafted and saliency-based mixing, breaking the  process into sub-tasks. Verma et al.~\citep{verma2019manifold} extended the Mixup concept to the model hidden layers, mixing feature maps instead of image pixels.

\noindent{\textbf{Generative Augmentation Methods:}} Though effective, the image-mixing paradigm faces an intrinsic limitation of ambiguous supervisory signal for the augmentation samples. Contemporary generative visual models can now generate remarkable high-quality synthetic samples~\citep{hoe2024interactdiffusion, qi2024deadiff, mahajan2024prompting, miao2024training}. Leveraging that, in self-supervised learning, StableRep \citep{tian2024stablerep} uses generative diffusion models for representation learning and augmentation, focusing on stable representations of real-world objects. Similarly, a popular work in supervised learning, DA-Fusion \citep{trabucco2023effective} directly uses generative and real image instead of parametric transformations to augment training data. Other works~\citep{trabucco2023effective, fu2024dreamda} have also demonstrated excellent potential of diffusion models in various applications generating diverse samples for data augmentation, also considering   foreground enhancement and background diversity for domain-specific concepts \citep{wang2024enhance}.
\par
\noindent{\bf Mixup with Generative Models:} To take advantage of both image-mixing and generative modeling, Islam et al.~\citep{islam2024diffusemix} recently proposed mixing images with their Image-to-Image generative  counterparts. However, the lose control over the generative content leads to low-quality augmentation samples in their approach - see Fig.~\ref{fig:comparison}. As compare to \citep{islam2024diffusemix}, we propose employing a Text-to-Image generative model where the generative content is explicitly tailored and filtered for semantic alignment with the original data. Moreover, our method also uses more sophisticated mixing mechanisms, enabled by the high-quality generative content of our method.

\section{Proposed Method}
\vspace{-1mm}
\noindent{\bf Overview:} 
Existing use of diffusion models in visual data augmentation relies on image-to-image (I2I) generation \citep{islam2024diffusemix, trabucco2023effective}, which lacks in control over the generative content. The central motivation of our technique is to enable a better control over the generative content to align it with the original data. We achieve this as the first approach that combines  text-to-image (T2I) generative modeling with the  image-mixing paradigm. 
\begin{wrapfigure}{r}{0.50\textwidth}
    \centering
    \includegraphics[width=0.48\textwidth]{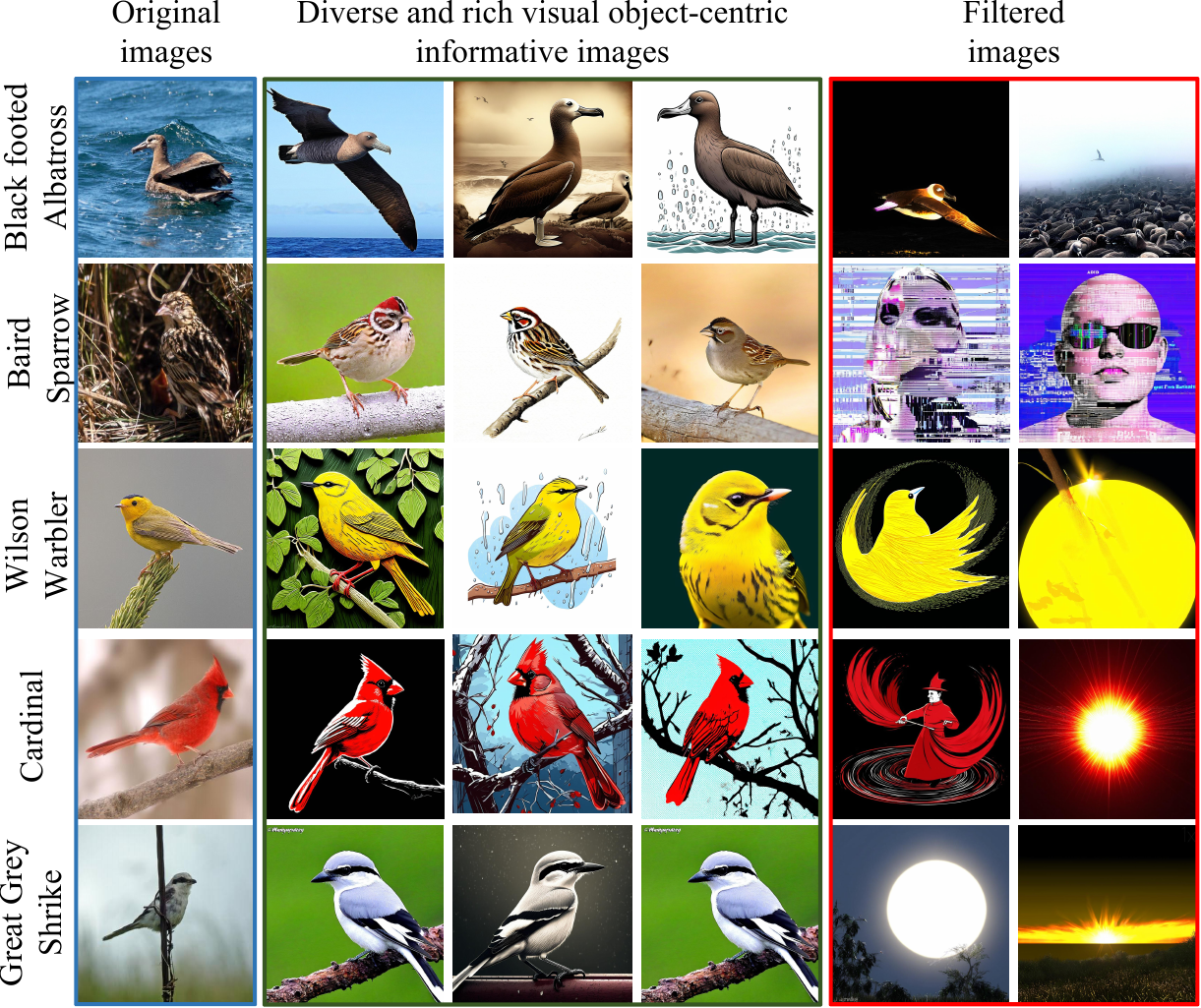}
    \captionsetup{width=0.48\textwidth}  
    \caption{Representative context guided generative images. Despite strong (positive and negative) context guidance, generated images may contain a small fraction ($\sim 10\%$ as confirmed by results in \S~\ref{sec:FAD}) of samples that semantically do not align well with the original images. 
    }
    \label{fig:filteration}
\end{wrapfigure}Figure~\ref{fig:overview}
provides an overview of the proposed method. Our T2I generative model is constrained with contextual and negative prompts to generate synthetic counterparts of a sample. Encoding the original and synthetic image pairs in CLIP encoder space~\citep{radford2021learning}, we estimate the semantic alignment of the generated images with the original sample. This is followed by filtering out the unaligned images and augmenting the data with patch- and pixel-level mixing with generated content.

\noindent{\bf Context Guidance for Generation: }
In \our{}, we propose to guide the generative process of a text-to-image (T2I) model with a combination of contextual prompt $\mathcal{P}_c$ and negative prompt $\mathcal{P}_n$. Herein, we term the collective guidance by $\mathcal{P}_c$ and $\mathcal{P}_n$ as contextual guidance. The standard forward diffusion process~\citep{rombach2022high, dhariwal2021diffusion}  adds noise to the input image \( x_0 \) in a step-by-step manner. This process is typically modeled as a Markov chain where noise is added at each step following the Gaussian distribution as 
\begin{equation}
p(x_t | x_{t-1}) = \mathcal{N}(x_t; \sqrt{\alpha_t} x_{t-1}, \beta_t {\bf I}),
\end{equation}
where \( x_t \) is the image at time step \( t \), \( \alpha_t \) controls the noise scale, ${\bf I}$ is the identity matrix and \( \beta_t \) is the variance of the added noise. Parameterized by a neural model, the reverse diffusion process denoises the sample,  progressively generating an image from noise. 
In \our{} the reverse process is conditioned on  $\mathcal{P}_c$ as well as $\mathcal{P}_n$. We let 
\begin{equation}
\begin{aligned}
p_\theta(x_{t-1} | x_t, \mathcal{P}_c, \mathcal{P}_n) &\propto 
\mathcal{N}\Big(x_{t-1}; \mu_\theta(x_t, \mathcal{P}_c, \mathcal{P}_n), \\
&\quad \Sigma_\theta(x_t, \mathcal{P}_c, \mathcal{P}_n)\Big),
\end{aligned}
\end{equation}
where \( \mu_\theta(x_t, \mathcal{P}_c, \mathcal{P}_n) \) and \( \Sigma_\theta(x_t, \mathcal{P}_c, \mathcal{P}_n) \) are the predicted mean and variance at step \( t \), \( \mathcal{P}_c \) guides the generated content toward the intended concept, and $\mathcal{P}_n$ specifies contents that should not be present in the generative output.

In a typical classifier-free T2I model, cross-attention \citep{chen2021crossvit} is used to provide context information to the reverse diffusion process. In that case, the predicted error of the image is computed as the difference between the conditional and unconditional error with a balancing term $\gamma$, i.e.,
\begin{equation}
   \tilde{\epsilon}_{\theta} = (1 + \gamma)\epsilon_{\theta}(x_t, \psi(q), t) - \gamma~\epsilon_{\theta}(x_t, \psi(\varnothing), t),
   \label{eq:orig}
\end{equation}
where $\epsilon_{\theta}$ is the approximation error, $\psi(q)$ computes the conditional signal for the text string $q$, and $\psi(\varnothing)$ denotes that the signal is computed by passing an empty string to the encoder. In our case, owing to the intended  objective of explicitly aligning the generative content with the original data semantics by  providing a stronger context, the predicted error is altered to the following 
\begin{equation}
   \tilde{\epsilon}_{\theta} = (1 + \gamma)\epsilon_{\theta}(x_t, \psi(\mathcal P_c), t) - \gamma~\epsilon_{\theta}(x_t, \psi(\mathcal P_n), t).
   \label{eq:NegPromp}
\end{equation}

Using the cross-attention, our T2I model is able to additionally leverage the negative prompt $\mathcal P_c$ along the (positive) prompt $\mathcal P_c$ to provide a comprehensive context guidance to the generation process. The intuition behind targeting Eq.~(\ref{eq:NegPromp}) in our method  comes from the insights of \citet{ban2024understanding} who affirm that the negative sign of the second term in Eq.~(\ref{eq:orig}) encourages removal of the content pertaining to the conditional signal from the generative output.

\noindent{\bf Hard Cosine Filtration:} 
Despite providing strong contextual guidance to the generative model, we observe semantic misalignment between the original samples and the generated outputs - see Fig.~\ref{fig:filteration}. Using misaligned images for data augmentation is detrimental.   
Hence, we devise a \textit{hard-cosine filtration}  to detect and ignore such undesired images. Let $\mathcal{I}, \mathcal{G} \subseteq \mathbb R^{h\times w \times c}$ be the sets of corresponding original and generated images, where $h,w$ and $c$ are height, width and channel dimensions. For the  filtration, we first compute the semantic  similarity $S(.,.)$ between the $i^{\text{th}}$ original sample $\mathcal I_i$ and its generative counterpart as
\[
S(\mathcal{I}_i, \mathcal{G}_i) = \frac{\psi_{\text{CLIP}}(\mathcal{I}_i) \cdot \psi_{\text{CLIP}}(\mathcal{G}_i)}{\|\psi_{\text{CLIP}}(\mathcal{I}_i)\| \|\psi_{\text{CLIP}}(\mathcal{G}_i)\|},
\]
where $\psi_{\text{CLIP}}$ denotes CLIP encoding of the image. Then, we let

\[
R(\mathcal{G}_i) = 
\begin{cases} 
1 & \text{if } S(\mathcal{I}_i, \mathcal{G}_i) > \tau, \\ 
0 & \text{otherwise}.
\end{cases}
\]
We retain \( \mathcal{G}_i \) if $R$ = 1 and discard it otherwise, subsequently generating another image and consider that as potential $\mathcal G_i$. Here, \( \tau \) is the threshold whose value is computed automatically, as discussed below. Let $\mathcal I^k \subset \mathcal I$ be a subset of the original images for the $k^{\text{th}}$ class, where $|\mathcal I^k| = N_k$. We can create $C^{N_{k}}_2$ pairs ($\mathcal I^k_i$, $\mathcal I^k_j$)$_{i\neq j}$ of these images. Since all these pairs are between real images, we can expect a measure for their semantic similarities $S(\mathcal{I}^k_i, \mathcal{I}^k_j)_{\forall(i,j)}$ to be a reliable handle over the semantics of the class data. Hence, we compute the Expected value of the similarity as 
\begin{equation}
\mathbb E[S(\mathcal{I}^k_i, \mathcal{I}^k_j)] = \frac{
\sum_{i = 1, j = 1}^{N_k} 
S(\mathcal I^k_i,\mathcal I^k_j)_{i\neq j}
}
{C^{N_{k}}_2},
\label{eq:expect}
\end{equation}
and let $\tau \approx \mathbb E[S(\mathcal{I}^k_i, \mathcal{I}^k_j)]$. Since $C^{N_{k}}_2$ becomes sizable even for mildly large class, e.g., 4,950 for $N_k = 100$, we approximate the Expectation values  by randomly selecting $z < C^{N_{k}}_2$ pairs and letting $N_k = z$ in Eq.~(\ref{eq:expect}).     

\vspace{0.5mm}
\noindent{\bf Image Mixing:} As a result of the above filtration we get a semantically well-aligned generated sample $\mathcal G_i$ for an original sample $\mathcal I_i$. We mix the two by employing a composite approach that considers both pixel-wise and patch-wise mixing. For the former, we use a mixing ratio variable $\lambda$ in the range $[0,1]$, and compute the resulting image as
\begin{equation}
\mathcal{M}^{\text{pixel}}_i = \lambda \mathcal{I}_i + (1 - \lambda) \mathcal{G}_i.
\label{eq:pix}
\end{equation}

For patch-wise mixing, our method can be interpreted as randomly cutting a patch from  $\mathcal{I}_i$ and paste it on  $\mathcal{G}_i$ or do the vice versa. Concretely, the mixed image is computed as
\begin{equation}
\mathcal{M}^{\text{patch}}_i = \mathbf{M}_p \odot \mathcal{I}_i + (1 - \mathbf{M}_p) \odot \mathcal{G}_i,
\label{eq:patch}
\end{equation}
where $\mathbf{M}_p$ is a binary mask of varying size indicating the patch to be cut from the image, $\odot$ represents element-wise multiplication. Finally, we model the selection of mixing for a given sample as a Bernoulli trial with success probability $\pi = 0.5$ due to two choices. For that, we sample $\eta \sim \text{Bernoulli}(\pi)$ and compute the mixed sample $\mathcal M_i$ as 
\begin{equation}
\mathcal{M}_i = 
\begin{cases} 
\mathcal{M}^{\text{pixel}}_i = \lambda \mathcal{I}_i + (1 - \lambda) \mathcal{G}_i, & \text{if } \eta = 1  \\
\mathcal{M}^{\text{patch}}_i = \mathbf{M}_p \odot \mathcal{I}_i + (1 - \mathbf{M}_p) \odot \mathcal{G}_i, & \text{if } \eta = 0
\end{cases}
\end{equation}

\section{Experiments}
\noindent{\textbf{Implementation Details.}} Our experiments are conducted using PyTorch on NVIDIA Tesla V100 and RTX 3090Ti GPUs, with training performed in both single-GPU and distributed data-parallel settings. The initial T2I prompt used to generate a contextual image was \textcolor{suppColor}{\texttt{Generate heavy snow to the <lab\_name>, a <dataset\_type> object}}. We augment data based on the number of images present in each class, utilizing Cosine-Continuous Stable Diffusion XL\footnote{https://huggingface.co/stabilityai/cosxl} in addition to the original one. The negative prompts varied depending on the dataset type.  We set the batch size to $16$, and models were trained for $300$ epochs using the SGD optimizer with a momentum of 0.9 and a weight decay of $5 \times 10^{-4}$. The initial learning rate was set to 0.01, which decayed by a factor of $0.1$ at epochs $150$ and $225$. 

\noindent{\textbf{Datasets.}} 
The used datasets are grouped into two categories of general and fined-grained classification tasks. For the general classification, we employ three popular datasets including {ImageNet-1K} \citep{imagenet} that contains diverse images, {CIFAR100} \citep{krizhevsky2009learning}; which is a $32$x$32$ image size dataset and {Tiny-ImageNet-200} \citep{le2015tiny} - a $64$x$64$ image size  dataset. For the {fine-grained image classification} category, we employ {Flower-102} \citep{nilsback2008automated} that contains $10$ images per class. We use this dataset for data scarcity. We also used  {Stanford Cars} \citep{krause20133d}, which  contains $196$ classes of different cars and models having different fined-grained details. We also use   {Birds-200-2011 (CUB)} \citep{wah2011caltech}, which  consists of $200$ classes of different bird species. These datasets cover a wide range of image distribution for a comprehensive evaluation.  

\begin{wraptable}{r}{0.50\textwidth}
    \captionsetup{width=0.50\textwidth}  
    \caption{Validation and testing set performance of the \our{} on Flower102 using PreAct-ResNet34 \textcolor{black}{backbone.}} 
    \centering
    \scalebox{0.70}[0.70]{
        \begin{tabular}{lcc}
            \toprule
            \rowcolor{mygray} Method & Valid Set (\%) & Test Set (\%) \\
            \midrule
            Mixup \citep{mixup} & 66.18 & 61.05 \\
            \arrayrulecolor{lightgray}\midrule
            CutMix \citep{cutmix} & 62.45 & 56.30 \\
            SaliencyMix \citep{uddin2020saliencymix} & 63.73 & 58.89 \\
            PuzzleMix \citep{kim2020puzzle} & 66.27 & 60.74 \\
            Co-Mixup \citep{kim2020co} & 65.10 & 59.02 \\
            Guided-AP \citep{kang2023guidedmixup} & 62.06 & 55.10 \\
            DiffuseMix \citep{islam2024diffusemix} & 67.28 & 60.82 \\
            \arrayrulecolor{black}\midrule
            \rowcolor{myblue} \textbf{\our{}}-\textbf{50} & \textbf{68.73}  & \textbf{ 61.07} \\
            \rowcolor{myblue} \textbf{\our{}}-\textbf{100}  & \textbf{69.84}  & \textbf{ 62.58} \\
            \arrayrulecolor{black}\bottomrule
        \end{tabular}
    }
    \label{tab:data_scarcity}
\end{wraptable} 

\noindent{\textbf{Baselines.}} To benchmark our method, we compare with SOTA methods that can be organized into three groups. (a)~Image-mixing methods \citep{kang2023guidedmixup, kim2020puzzle, cutmix, mixup, uddin2020saliencymix, huang2021snapmix, hendrycksaugmix, verma2019manifold}. These are SOTA  methods that mix source and target images in  pixel-, patch- or saliency-wise manner. (b) Generative methods \citep{hesynthetic, trabucco2023effective, wang2024enhance}. Instead of using basic image transformations to construct augmentation samples, they generate samples for data augmentation using generative models. (c)~A method using image-mixing with generative content \citep{islam2024diffusemix}. Conceptually, this method is closest to our approach.

\begin{table*}
    \centering 
    \scalebox{0.75}{
            \begin{tabular}{p{5.5cm}p{1.3cm}<{\centering}p{1.5cm}<{\centering}|p{1.5cm}<{\centering}p{1.5cm}<{\centering}|p{1.5cm}<{\centering}p{1.5cm}<{\centering}}
                \toprule
                \rowcolor{mygray} 
                Method & \multicolumn{2}{c|}{ImageNet-1K} & \multicolumn{2}{c|}{Tiny ImageNet-200} & \multicolumn{2}{c}{CIFAR-100} \\ 
                \rowcolor{mygray} 
                & \begin{tabular}[c]{@{}c@{}}Top-1 (\%) \end{tabular} & \begin{tabular}[c]{@{}c@{}}Top-5 (\%) \end{tabular} 
                & \begin{tabular}[c]{@{}c@{}}Top-1 (\%) \end{tabular} & \begin{tabular}[c]{@{}c@{}}Top-5 (\%) \end{tabular} 
                & \begin{tabular}[c]{@{}c@{}}Top-1 (\%) \end{tabular} & \begin{tabular}[c]{@{}c@{}}Top-5 (\%) \end{tabular} \\ 
                \midrule
                Mixup \citep{mixup} & 77.03 & 93.52 & 56.59 & 73.02 & 76.84 & 92.42 \\
                \midrule
                AugMix \citep{hendrycksaugmix} & 76.75 & 93.30 & 55.97 & 74.68 & 75.31 & 91.62 \\
                Manifold Mixup \citep{verma2019manifold} & 76.85 & 93.50 & 58.01 & 74.12 & 79.02 & 93.37 \\
                CutMix \citep{cutmix} & 77.08 & 93.45 & 56.67 & 75.52 & 76.80 & 91.91 \\
                PixMix \citep{hendrycks2022pixmix} & 77.40 & - & - & - & 79.70 & - \\
                PuzzleMix \citep{kim2020puzzle} & 77.51 & 93.76 & 63.48 & 75.52 & 80.38 & 94.15 \\
                GuidedMixup \citep{kang2023guidedmixup} & 77.53 & 93.86 & 64.63 & 82.49 & 81.20 & 94.88 \\
                Co-Mixup \citep{kim2020co} & 77.63 & 93.84 & 64.15 & - & 80.15 & - \\
                DiffuseMix \citep{islam2024diffusemix} & 78.64 & 95.32 & 65.77 & 83.66 & 82.50 & 95.41 \\
                \midrule
                \rowcolor{myblue} \textbf{\our{}-50} & \textbf{79.47} & \textbf{96.32} & \textbf{65.91} & \textbf{84.24} & \textbf{82.84} & \textbf{96.24} \\ 
                \rowcolor{myblue} \textbf{\our{}-100} & \textbf{80.48} & \textbf{98.21} & \textbf{67.81} & \textbf{87.38} & \textbf{83.37} & \textbf{97.62} \\
                \bottomrule
            \end{tabular}
            }
            \caption{Comparison of  Top-1  and Top-5  classification performance  across three popular datasets ImageNet-1K, Tiny ImageNet-200, and CIFAR-100 using different data augmentation techniques. 
            For ImageNet-1K, ResNet-50 is used as the backbone, while PreActResNet-18 is employed for Tiny ImageNet-200 and CIFAR-100. *-50 and *-100 variants of our method respectively augment 50\% and 100\% training data. }
            \label{tab:general_class}
\end{table*}

\section{State-of-the-art Comparison}

\noindent{\textbf{General Classification:}} General Classification (GC) serves as a critical benchmarking task to assess the impact of data augmentation techniques on the models \citep{dosovitskiy2020image, resnet}. For GC, we evaluate \our{} on ImageNet-1K \citep{imagenet}, Tiny-ImageNet-200 \citep{le2015tiny} and CIFAR-100 \citep{krizhevsky2009learning}, highlighting the performance gains over the  existing  methods. Following \citep{islam2024diffusemix, kang2023guidedmixup}, we trained ResNet-50 on ImageNet-1K with two variants. \our{}-50 augments \numbersBlue{$50\%$} training data and \our{}-100 augments \numbersBlue{$100\%$} training  data. The results are summarized in Table~\ref{tab:general_class}.

\begin{wraptable}{r}{0.50\textwidth}
    \captionsetup{width=0.50\textwidth}  
    \caption{Top-1 (\%) performance comparison for the  fine-grained visual categorization task using DenseNet121.}
    \scalebox{0.75}[0.75]{
    \begin{tabular}{lccc}
    \toprule
    \rowcolor{mygray} 
    Method & CUB & Cars & Caltech \\\midrule
    Mixup \citep{mixup} & 74.23 & 89.06 & 91.47 \\
    \arrayrulecolor{lightgray}\midrule
    CutMix ~\citep{cutmix} & 74.30 & 88.84 & 91.36 \\
    SaliencyMix \citep{uddin2020saliencymix} & 68.75 & 88.91 & 90.78 \\
    PuzzleMix~\citep{kim2020puzzle} & 77.27 & 90.10 & 91.47 \\
    Co-Mixup~\citep{kim2020co} & 77.05 & 90.23 & 90.44 \\
    Guided-AP~\citep{kang2023guidedmixup} & 77.52 & 90.23 & 91.84 \\
    DiffuseMix~\citep{islam2024diffusemix} & 77.82 & 90.83 & 92.03 \\
    \arrayrulecolor{black}\midrule
    \rowcolor{myblue} \textbf{\our{}-50} & \textbf{78.12} & \textbf{91.17} & \textbf{92.23} \\
    \rowcolor{myblue} \textbf{\our{}-100} & \textbf{79.52} & \textbf{92.71} & \textbf{93.72} \\
    \arrayrulecolor{black}\bottomrule
    \end{tabular}}
    \label{tab:fgvc_acc}
\end{wraptable}

For Top-1 and Top-5 accuracies, our both variants clearly outperform all previous methods. On ImageNet-1K, \our{}-100 achieves a very strong performance, outperforming Co-Mixup - a popular saliency-guided image-mixing method - by nearly \numbersBlue{$3\%$} in Top-1 accuracy and by nearly  \numbersBlue{$5\%$} in Top-5 accuracy. For the \our{}-50 variant, the gains are relatively low, which is intuitive. However, we still outperform DiffuseMix~\citep{islam2024diffusemix} by an absolute \numbersBlue{$1\%$} and \numbersBlue{$0.83\%$} for Top-5 and Top-1 accuracies.

On Tiny ImageNet-200, as compared to the second best performer \citep{islam2024diffusemix},  \our{}-100 gains an absolute improvement of \numbersBlue{$2.04\%$} for Top-1 accuracy and \numbersBlue{$3.72\%$} for Top-5 accuracy. The trends remain for the CIFAR-100 dataset. As  compared to the popular Mixup baseline \citep{mixup} which also works on image-mixing principles, our notable absolute performance gains for Tiny ImageNet-200 are  \numbersBlue{$11.22\%$} and \numbersBlue{$14.36\%$}. This highlights the benefit of using T2I models for image mixing. 

\begin{wraptable}{r}{0.50\textwidth}
    \captionsetup{width=0.50\textwidth} 
    \caption{Top-1 (\%) performance of \our{} on fine-tuning experiments using ImageNet pretrained Wide ResNet-101.}
    \scalebox{0.75}[0.75]{
    \begin{tabular}{lcc}
    \toprule
    \rowcolor{mygray} 
    Method & CUB-Birds & Flowers-102  \\ \midrule
    ResNet-50 & 78.61 & 88.26 \\
    \arrayrulecolor{lightgray}\midrule
    Mixup \citep{mixup} & 79.37 & 89.63  \\
    CutMix ~\citep{cutmix} & 79.42  & 90.84  \\
    SaliencyMix \citep{uddin2020saliencymix} &79.73 & 91.43  \\
    SnapMix~\citep{huang2021snapmix} & 79.80 & 91.64  \\
    DiffuseMix~\citep{islam2024diffusemix} & 80.23 & 93.45  \\
    \arrayrulecolor{black}\midrule
    \rowcolor{myblue} \textbf{\our{}-50} & \textbf{81.57} & \textbf{94.38}  \\
    \rowcolor{myblue} \textbf{\our{}-100} & \textbf{82.37} & \textbf{95.10}  \\
    \arrayrulecolor{black}\bottomrule
    \end{tabular}}
    \label{tab:transfer_learning}
\end{wraptable}

\noindent{\textbf{Fined-Grained Visual Classification:}} 
Fine-grained visual categorization (FGVC)  tasks, such as distinguishing between fined-grained categories where intra-class is similar, are inherently challenging due to the subtleness of differences between the categories \citep{wei2023fine,  tang2023weakly}. 
They demand models capable of extracting indistinguishable features, requiring sophisticated data augmentation techniques to enhance generalization performance \citep{islam2024diffusemix}. In Table~\ref{tab:fgvc_acc}, we summarize the results for FGVC tasks on three datasets. \our{} leads to across the board  improvements  over the SOTA methods. On the CUB dataset, known for its demanding task of bird species recognition, \our{} is able to push the performance to an absolute \numbersBlue{$1.7\%$}, showcasing strong ability to capture fine-grained details. For the Cars dataset, an absolute gain of \numbersBlue{$1.88\%$} over SOTA and  \numbersBlue{$3.06\%$} over the popular baseline \citep{mixup} is visible.Finally, on Caltech, which encompasses a broader range of object categories, our method  continues to outperform the existing method in a similar fashion.

\noindent{\textbf{Data Scarcity:}} Data scarcity is a  prevalent challenge for deep learning, especially in the  domains pertaining to fine-grained visual classification  where data labeling is costly ~\citep{nilsback2008automated}. GuidedMixup \citep{kang2023guidedmixup} is among the first augmentation methods to report performance  for the data scarcity problem using  paring algorithm. Islam et al.~\citep{islam2024diffusemix} followed \citep{kang2023guidedmixup} in their comprehensive benchmarking.
We also compare \our{} with the  SOTA methods, e.g.,  \citep{islam2024diffusemix}, \citep{kang2023guidedmixup},  \citep{mixup} under a limited data regime on the Flower102 dataset, using $10$ images per class. As summarized in Table \ref{tab:data_scarcity}, \our{} achieves high  performance gains for this problem. As compared to the second best performer DiffuseMix \citep{islam2024diffusemix},
our \our{}-100 achieves  absolute performance gains of $\sim$ \numbersBlue{$2.56\%$} and $\sim$ \numbersBlue{$1.76\%$} 
on the validation and test sets respectively. Even our DiffCoRe-Mix-50 variant comprehensively outperforms all existing methods.

\begin{wrapfigure}{r}{0.55\textwidth}
    \centering
    \includegraphics[width=0.55\textwidth]{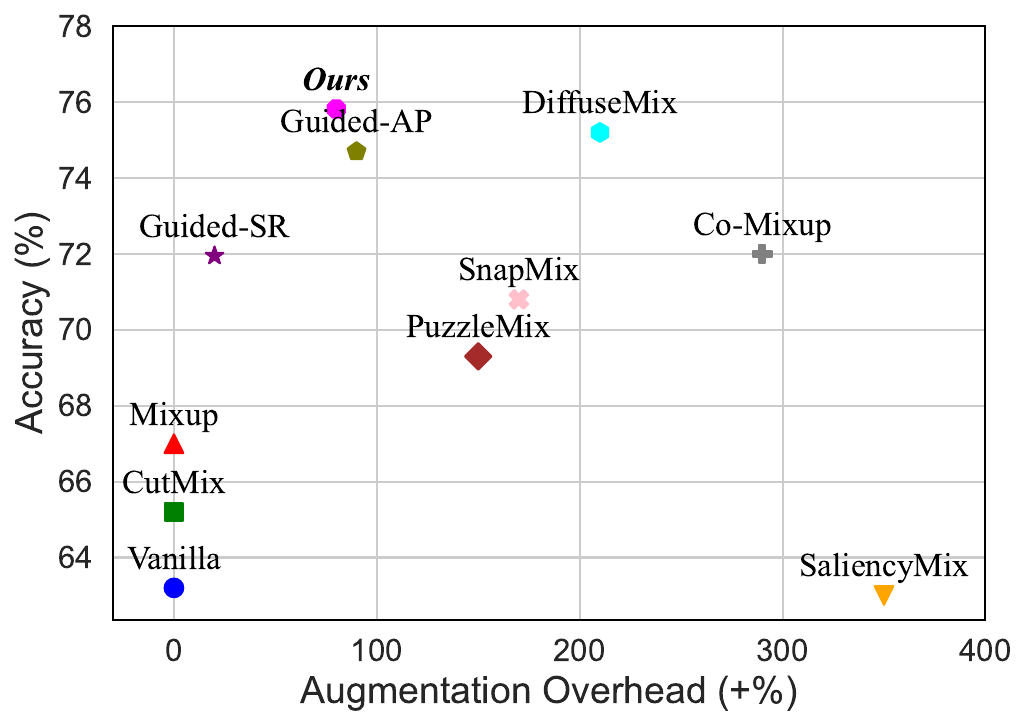}
    \caption{Augmentation overhead (+\%) - accuracy (\%) plot
    on CUB dataset with batch size 16. The closer the value to the upper left corner, the better the augmentation \textcolor{black}{strategy.}} 
    \label{fig:aug_overhead}
\end{wrapfigure}

\noindent{\textbf{Transfer Learning:}} Transfer learning enables leveraging pre-trained weights to improve model performance on new datasets with relatively smaller training sizes \citep{islam2024diffusemix, kang2023guidedmixup}. In our transfer learning experiments, we fine-tuned a Wide ResNet-101 model pretrained on ImageNet-1K \citep{imagenet} to evaluate the effectiveness of various data augmentation methods. The results are summarized in Table~\ref{tab:transfer_learning}.

Comparing \our{} with the best performing exiting method  DiffuseMix \citep{islam2024diffusemix}, we observed notable gains. On the CUB-Birds dataset, \our{}-50 achieves an accuracy gain of \numbersBlue{$1.34\%$}.  \our{}-100 further improves the performance to \numbersBlue{$82.37\%$}, yielding a gain of \numbersBlue{$2.14\%$} over DiffuseMix. On the Flowers-102 dataset, \our{}-50 outperforms DiffuseMix by \numbersBlue{$0.93\%$} \numbersBlue{$(94.38\%$} vs. \numbersBlue{$93.45\%$}), and \our{}-100 extends this margin to \numbersBlue{$1.65\%$}, reaching a Top-1 accuracy of \numbersBlue{$95.10\%$}. These consistent improvements show that our proposed method significantly enhances the fine-tuning performance compared to the baseline approaches.

\begin{wraptable}{r}{0.50\textwidth}
    \captionsetup{width=0.50\textwidth} 
\centering
\caption{Augmentation overhead (\%) for different methods with varying batch sizes.}

\scalebox{0.8}{
\begin{tabular}{l>{\centering\arraybackslash}p{1.1cm}>{\centering\arraybackslash}p{1.1cm}>{\centering\arraybackslash}p{1.55cm}}
\toprule
\rowcolor{mygray}
Method & \multicolumn{3}{c}{Augmentation Overhead (+\%)} \\
\rowcolor{mygray}
 & 16 & 32 & 64 \\
\midrule
Mixup & 0.9 & 0.6 & 0.4 \\
\arrayrulecolor{lightgray}\midrule
CutMix & 1.5 & 1.0 & 0.6 \\
SaliencyMix & 353.3 & 701.8 & 923.3 \\

SnapMix & 67.4 & 64.9 & 60.2 \\
PuzzleMix & 138.5 & 139.9 & 134.1 \\
Co-Mixup & 292.1 & 490.2 & 716.6 \\
Guided-AP (Random) & 87.8 & 81.9 & 70.1 \\
Guided-AP (Greedy) & 89.2 & 83.0 & 77.5 \\
\arrayrulecolor{black}\midrule
\rowcolor{myblue} \textbf{ \our{}} & \textbf{68.2} & \textbf{59.3} & \textbf{58.9} \\
\arrayrulecolor{black}\bottomrule
\end{tabular}
}
\label{tab:aug_overhead}
\end{wraptable}

\noindent{\textbf{Computational Overhead:}} 
We analyze the computational overhead against the performance gain of our method and compare it with that of the  SOTA methods in Fig.~\ref{fig:aug_overhead}. 
Following  \citep{islam2024diffusemix} and  \citep{kang2023guidedmixup}, we define the overhead $\mathcal{A_{O}}$ as:
\begin{equation}
\mathcal{A_{O}} = \frac{\mathcal{T}_{aug} - \mathcal{T}_{van}}{\mathcal{T}_{van}} \times 100 (\%),
\end{equation}
where $\mathcal{T}_{aug}$ denotes the  training time after image generation, and $\mathcal{T}_{van}$ is the training time of the baseline model \citep{resnet}  without augmentation. We generate images beforehand and use them  for the rest of the training. \our{} demonstrates a remarkable trade-off between performance and augmentation overhead, outperforming all other methods in accuracy while keeping overhead significantly lower than Co-Mixup \citep{kim2020co} and SaliencyMix \citep{uddin2020saliencymix}. We also report the training overhead with varying batch sizes in Table~\ref{tab:aug_overhead}. As the results show, our method retains one of the lower overheads. This is in addition to the fact that {\our} achieves considerable performance gains across the board.

\begin{wrapfigure}{r}{0.60\textwidth}
    \centering
    \includegraphics[width=0.60\textwidth]{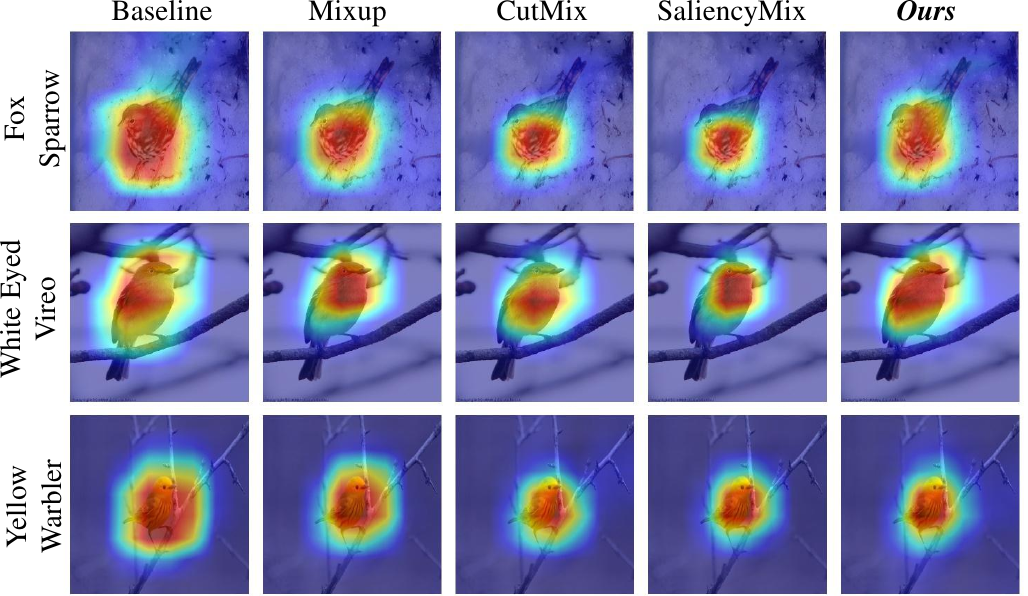}
    \caption{\textbf{Representative saliency visualizations} on original data samples. Our method guides the model to more precisely focus on the target object in \textcolor{black}{the image}. } 
    \label{fig:gradcam}
\end{wrapfigure}

\noindent{\textbf{Saliency Visualization:}} To evaluate the impact of augmentation on model's attention in the decision-making process, we compare the saliency maps for \our{} with the  popular data augmentation methods \citep{resnet, uddin2020saliencymix, cutmix, mixup}. As shown in Fig.~\ref{fig:gradcam},  the typical saliency maps for \our{} appear more consistently around the regions of foreground object. The outputs show more concentrated activation around salient regions of the birds, especially around their distinguishing features like the head or unique color patterns. This indicates that \our{} effectively emphasizes on the key features while encouraging the network to learn discriminative regions under diverse augmentations. \our{} offers a balanced approach, enhancing robustness while preserving key discriminative features, making it suitable for scenarios demanding a balance between precision and generalization.

\begin{figure*}[t]
    \centering
        \includegraphics[width=1.0\linewidth]{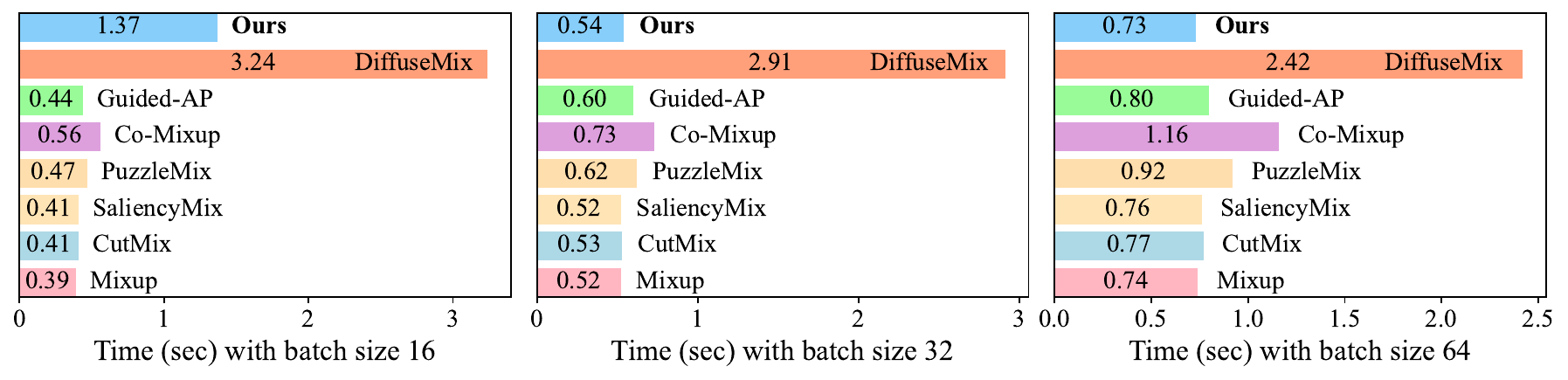}
    \caption{
    Comparison of batch processing time (sec.) on  CUB dataset. Batch sizes of $16$, $32$, and $64$ are used on the \textcolor{black}{same hardware}. 
    } 
    \label{fig:batch_time}
\end{figure*}

\noindent{\textbf{Batch Processing  Time:}} We also compare \our{} with other data augmentations methods \citep{mixup, cutmix, uddin2020saliencymix, kim2020co, kang2023guidedmixup, islam2024diffusemix} in terms of batch processing time. Results are summarized in Fig.~\ref{fig:batch_time}. For a batch size of $16$, Mixup demonstrates a faster training time of \numbersBlue{$0.39$} seconds compared to \our{} \numbersBlue{$1.37$} seconds, indicating that mixup is  more efficient for small batch training.

However, as the batch size increases to $32$, the time difference narrows, with Mixup at \numbersBlue{$0.52$} seconds and \our{} at \numbersBlue{$0.54$} seconds. This convergence suggests that our method scales more effectively with batch size compared to Mixup. The pattern continues with a batch size of $64$, where Mixup records \numbersBlue{$0.74$} seconds, while \our{} is slightly faster at \numbersBlue{$0.73$} seconds. 

\section{Further Analysis and Discussions}
\label{sec:FAD}

We provide a detailed ablation study to evaluate important aspects of our proposed method and design choices.

\noindent \textbf{Ablation Studies:} We investigate the efficacy of \emph{pixel-wise} and \emph{patch-wise} design choices in our \our{}. To show the effectiveness of our selection, we also report the individual  performance of the two variants. All ablation studies are conducted using ResNet-50 \citep{resnet} backbone. In Table \ref{tab:our_ablation}, we first use ResNet-50  as the baseline, which is solely trained on traditional augmentation. 
It achieves a Top-1 accuracy of \numbersBlue{$85.86\%$} and a Top-5 accuracy of \numbersBlue{$91.19\%$} on the Stanford Cars dataset. We observe that introducing our Generative Augmentation (GenAug) alone brings \numbersBlue{$1.82\%$}  and \numbersBlue{$5.60\%$} absolute  gains for Top-1 and Top-5 accuracies.

\begin{wraptable}{r}{0.5\textwidth}
    \captionsetup{width=0.5\textwidth}
    \caption{Performance comparison  on Stanford Cars dataset.}

    \scalebox{0.8}{
    \begin{tabular}{lcc}
        \toprule
        \rowcolor{mygray} 
        Mixing Strategies & Top-1 (\%) & Top-5 (\%) \\
        \midrule
        ResNet-50 
        \citep{resnet} & 85.86 & 91.19 \\
        \arrayrulecolor{lightgray}\midrule
        R50 + GenAug   & 87.68 & 96.79 \\
        R50 + Pixel-wise & 88.08 & 97.15 \\
        R50 + Patch-wise  & 88.92 & 97.73 \\
        R50 + GenAug + Pixel-wise     & 89.50 & 97.86 \\
        R50 + GenAug + Patch-wise    & 90.62 & 98.49 \\
        \arrayrulecolor{black}\midrule
        \rowcolor{myblue} 
        \textbf{\our{}-50} & \textbf{91.85} & \textbf{98.83} \\
        \rowcolor{myblue} 
        \textbf{\our{}-100} & \textbf{92.74} & \textbf{99.46} \\
        \bottomrule
    \end{tabular}
    }
    \label{tab:our_ablation}
\end{wraptable}

These results are consistent with the findings in \citep{wang2024enhance, trabucco2023effective}. 
Further, we individually examine the pixel-wise and patch-wise technique. By adding the pixel-wise mixing, the Top-1 accuracy increases to \numbersBlue{$88.08\%$} and the Top-5 accuracy to \numbersBlue{$97.15\%$}, highlighting the effectiveness of pixel-wise augmentation in improving model generalization. Similarly, in our  experiment with the patch-wise approach, we again observe a slight performance improvement, where Top-1 accuracy increases to \numbersBlue{$88.92\%$} and the Top-5 accuracy to \numbersBlue{$97.73\%$}. Combining GenAug with pixel-wise mixing further improves the performance. 
This suggests a synergistic effect 
A similar trend is visible for the patch-wise mixing with  GenAug. 

In our eventual approach, we use  two variants of \our{}-50 (which uses 50\% generative augmentation along with original data) and \our{}-100  (which utilizes 100\% generative augmentation with real data). They incorporate both pixel-wise and patch-wise mixing. They achieve impressive performances with a Top-1 accuracy of \numbersBlue{$91.85\%$} and a Top-5 accuracy of \numbersBlue{$98.83\%$} for \our{}-50. Our \our{}-100 shows even stronger performance, achieving further absolute gains of  \numbersBlue{$0.89\%$} and \numbersBlue{$0.63\%$} for Top-1 and Top-5 performances. 

\begin{wrapfigure}{r}{0.5\textwidth}
    \centering
    \includegraphics[width=0.5\textwidth]{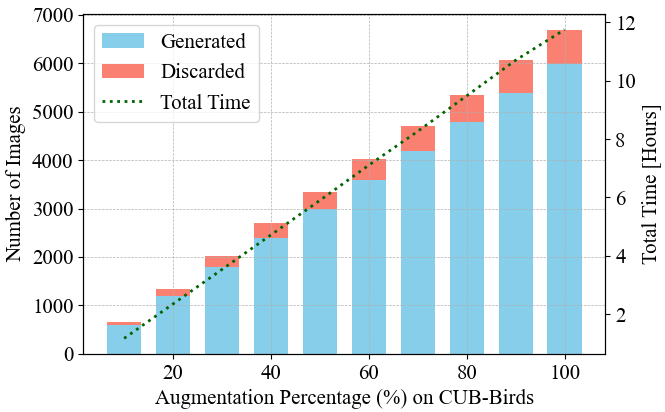}
    \caption{\textcolor{black}{Ablation studies on the percentage of generative augmentation via CosXL.}}
    \label{fig:inference_gpu}
\end{wrapfigure}

\noindent\textbf{Generative Inference Computational Cost:} 
Our method allows generating augmentation samples in varying percentages of the original data. 
\textcolor{black}{Figure \ref{fig:inference_gpu}},  summarizes the computational overhead associated with the generative inference for \textcolor{black}{CUB-Birds} dataset, where we vary the percentages of $512$x$512$ augmentation samples from \numbersBlue{$10$} to \numbersBlue{$100$}. 

We also report the percentage of images automatically discarded by our filtration method.  It can be observed that this percentage remains consistent in the range \numbersBlue{$[10.1, 11.1]$}. This ensures a linear generative inference complexity incurred by our technique, which is desirable for scalability. Our total compute timings (hours) show that even 100\% augmentation is fully feasible for a small dataset.

 \section{Conclusion}
 \vspace{-1mm}
 In this work, we proposed \our{}, a reliable T2I based data augmentation approach that employs diffusion models and various prompts to generate domain-specific class-relevant samples to increase diversity in the training dataset. Our method is intended to provide augmentation samples that are responsibly generated to align with the training data. Our technique mixes real and generative images following a systematic approach that considers both patch and pixel level mixing. On multiple tasks; such as fine-grained classification, general classification, data scarcity, finetuning, and augmentation overhead, we demonstrate notable performance gains on several benchmark datasets including ImageNet-1K, StanfordCars, Tiny-ImageNet-200, CIFAR-100, Flower-102, Caltech Birds. We also demonstrate string computational efficacy for large training batches. Moreover, our result show that the our augmentation samples lead to more precise saliency maps for the induced models. 

\subsubsection*{Acknowledgments}
Dr.~Naveed Akhtar is a recipient of the Australian Research Council Discovery Early Career Researcher Award (project number DE230101058) funded by the Australian Government. This work is also partially supported by Google Research under Google Research Scholar Program.

\bibliography{iclr2025_conference}
\bibliographystyle{iclr2025_conference}

\newpage
\appendix
\section{Appendix}
This supplementary document contains additional results presented in the order of mention in the main paper. 

\begin{itemize}
    
    
    \item Section \ref{supp:batch_time} provides further  study of generative inference computational cost across Stanford Cars and Flower102.

    \item Section \ref{lab:vit_ours} provides additional experiments for  SOTA vision transformer models such as FastViT, MobileViT and EfficientViT.
    
    \item Section \ref{supp:diversity} visually compare the  fidelity and diversity of \our{} with DiffuseMix \cite{islam2024diffusemix}, Real-Guidance \cite{hesynthetic}, DA-Fusion \cite{trabucco2023effective}, and Diff-Mix \cite{wang2024enhance}.

    
    \item Section \ref{supp:batch_visulization} provides training batch visualization comparison of data augmentation strategies including Mixup \cite{mixup}, CutMix \cite{cutmix}, SaliencyMix \cite{uddin2020saliencymix}, PuzzleMix \cite{kim2020puzzle}, GuidedMixup \cite{kang2023guidedmixup} and our proposed method.
\end{itemize}

\section{Generative Inference Computational Cost}
\label{supp:batch_time}

\begin{figure*}
    \centering
    \begin{subfigure}{0.47\linewidth}
        \centering
        \includegraphics[width=\linewidth]{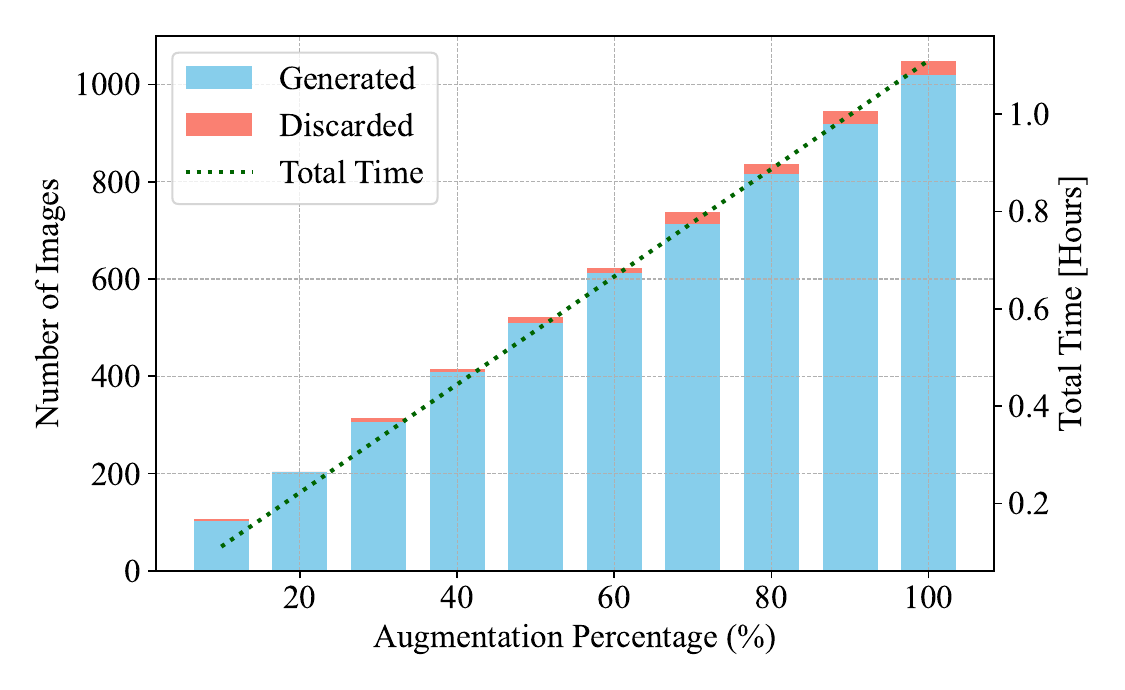}
        \caption{Computational cost on Flowers102 dataset}
        \label{fig:infer_flower}
    \end{subfigure}
    \begin{subfigure}{0.47\linewidth}
        \centering
        \includegraphics[width=\linewidth]{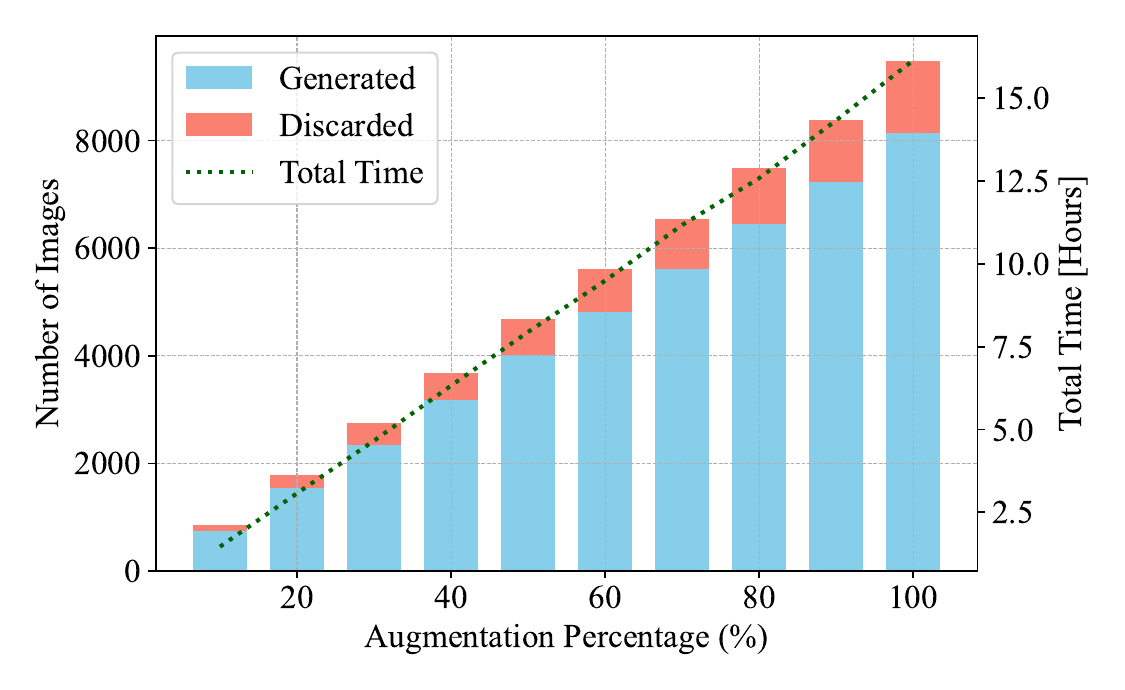}
        \caption{Computational cost on Stanford Car dataset}
        \label{fig:infer_car}
    \end{subfigure}
    \caption{Overall computational cost analysis on two fined-grained datasets.}
    \label{fig:compuational_cost}
\end{figure*}

To further analyze the computational cost, we extend this study on Flower102 and Stanford Cars dataset.
\par
In data scarcity scenarios, Flowers-102 is a popular dataset because it contains $10$ images per class. As presented in Fig. \ref{fig:infer_flower}, we increased the augmentation percentage from \numbersBlue{$10\%$} to \numbersBlue{$100\%$}, there was a incremental growth in the number of generated images. We began with $102$ images at \numbersBlue{$10\%$} augmentation and peaked at \numbersBlue{$1020$} images at \numbersBlue{$100\%$}, demonstrating that higher augmentation levels typically lead to greater output. Furthermore, to examine the quality of generated images, we encountered a notable number of discarded images, particularly at higher augmentation levels. For instance, at \numbersBlue{$70\%$}, we generated \numbersBlue{$714$} images but discarded \numbersBlue{$23$}, indicating that while the quantity rose, quality control became increasingly challenging. Thus, our hard-cosine filtration approach is important  to ensure semantically aligned images with training data. In terms of generation time, we note that generation time increased significantly with higher augmentation percentages. However, the trend is linear because the percentage of the discarded samples remains steady. 
\par
The similar trend continues for the Stanford Cars datasets (see in Fig. \ref{fig:infer_car}). The computational demands and efficiency of the CosXL T2I model as generative augmentation increases for the Stanford Car dataset. Notably, inference time rises proportionally with augmentation level, revealing the significant cost associated with higher generative percentages. At \numbersBlue{$10\%$}, the model generated \numbersBlue{$734$} images in under \numbersBlue{$1.5$} hours, but at \numbersBlue{$100\%$}, it took over \numbersBlue{$16$} hours to generate \numbersBlue{$8,144$} images. Discard rates, which hover around \numbersBlue{$15\%$} across levels, suggest that quality control remains consistent regardless of augmentation percentage. The incremental time increase emphasizes the trade-off between quantity and inference speed. While higher augmentation enhances dataset diversity, it also heavily impacts on generation time. 

\begin{wraptable}{r}{0.50\textwidth}
    \captionsetup{width=0.50\textwidth}
    \caption{Top-1 (\%) performance of \our{} on different vision transformer methods.}

    \scalebox{0.85}[0.85]{
    \begin{tabular}{lcc}
    \toprule
    \rowcolor{mygray} 
    Method & CUB-Birds & Flowers-102  \\ \midrule
    
    MobileViT  & 54.73 & 45.31  \\
    \rowcolor{myblue} +\textbf{\our{}-100} & \textbf{67.32} & \textbf{53.45} \\
    \arrayrulecolor{lightgray}\midrule
    
    FastViT  & 51.63  & 47.10  \\
    \rowcolor{myblue} + \textbf{\our{}-100} & \textbf{61.13} & \textbf{48.41} \\
    \arrayrulecolor{lightgray}\midrule
    
    EfficientViT & 42.65 & 37.15  \\
    \rowcolor{myblue} + \textbf{\our{}-100} & \textbf{50.53} & \textbf{49.90} \\
    
    \arrayrulecolor{black}\bottomrule
    \end{tabular}}
    \label{tab:viT}
\end{wraptable}

\section{\our{} with ViT Methods}
\label{lab:vit_ours}
We further extend our experiments on recent SOTA Vision Transformer (ViT) \cite{mehtamobilevit, woo2023convnext, vasu2023fastvit} methods to observe the compatibility of \our{} with vision transformers. For this, we pick pure ViTs and convolutional ViTs methods to access the performance. The motivation behind these experiments is to see how ViTs methods perform when the class contain real and generative data. 
\par
\begin{figure*}[t]
    \centering
    \includegraphics[width=1\linewidth]{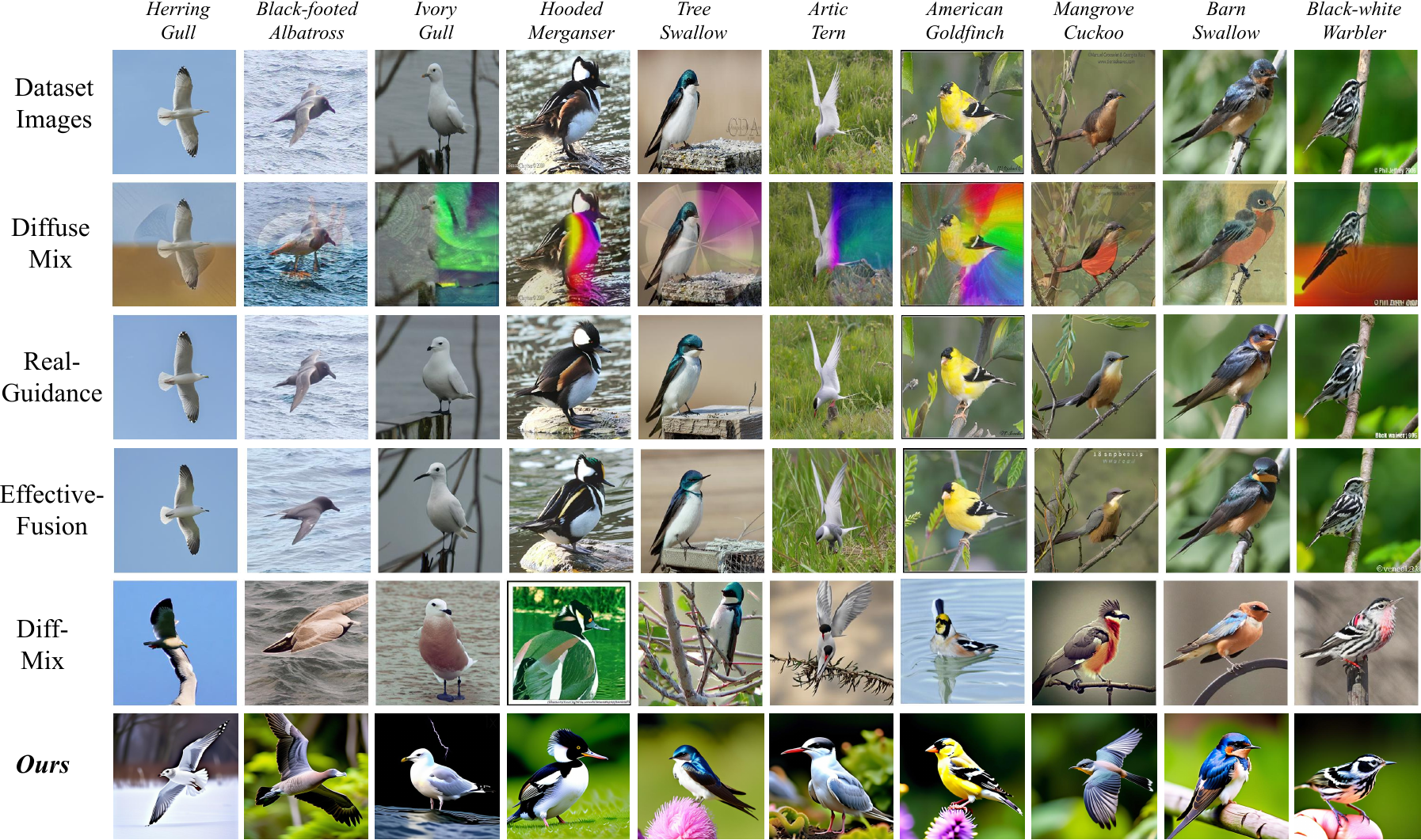}
    \caption{Diverse visual examples of recent generative data augmentation methods such as DiffuseMix \cite{islam2024diffusemix}, Real-Guidance \cite{hesynthetic}, Effective-Fusion \cite{trabucco2023effective}, Diff-Mix \cite{wang2024enhance} and Ours. The original image taken from paper \cite{anonymous2024decoupled} and then we added our method in this image. Our method shows strong semantic alignment with natural images along high quality of the outputs.}
    \label{fig:diversity_sota}
\end{figure*}

Our method, \our{}-100, in Table \ref{tab:viT} demonstrates substantial improvements across various vision transformer models in terms of Top-1 accuracy. Notably, when applied to MobileViT \cite{mehtamobilevit}, \our{}-100 achieves an impressive gain of \numbersBlue{$12.59\%$} over the baseline, elevating accuracy from \numbersBlue{$54.73\%$} to \numbersBlue{$67.32\%$} on the CUB-Birds dataset. Similarly, on the Flowers-102 dataset, \our{}-100 enhances MobileViT performance by \numbersBlue{$8.14\%$}, increasing the accuracy from \numbersBlue{$45.31\%$} to \numbersBlue{$53.45\%$}. For FastViT \cite{vasu2023fastvit}, the introduction of \our{}-100 results in a \numbersBlue{$9.50\%$} boost, pushing accuracy from \numbersBlue{$51.63\%$} to \numbersBlue{$61.13\%$} on CUB-Birds, while showing a moderate improvement of \numbersBlue{$1.31\%$} on Flowers-102, from \numbersBlue{$47.10\%$} to \numbersBlue{$48.41\%$}. Additionally, \our{}-100 significantly raises EfficientViT \cite{liu2023efficientvit} accuracy on Flowers-102 by \numbersBlue{$12.75\%$}, transforming a baseline of \numbersBlue{$37.15\%$} to \numbersBlue{$49.90\%$}, and provides a notable uplift of \numbersBlue{$7.88\%$} on CUB-Birds, from \numbersBlue{$42.65\%$} to \numbersBlue{$50.53\%$}. These results consistently underscore the effectiveness of \our{}-100 across diverse architectures, positioning it as a robust enhancement for vision transformer performance.


\section{Diversity and Fidelity}
\label{supp:diversity}
We compare the quality of generated images obtained from our contextual prompts while considering negative prompts to evaluate diversity and fidelity of our approach. We visually compare \our{} with other SOTA methods. The results indicates that \our{} have introduced substantial diversity and color variations in class-relevant objects as compare to other methods \cite{islam2024diffusemix, wang2024enhance, hesynthetic, trabucco2023effective}. Figure~\ref{supp:diversity} shows examples  of different SOTA generative data augmentation methods, validating the reason of performance gains. While comparing, we observe that DiffuseMix \cite{islam2024diffusemix} diversifies half input image from a stylistic perspective rather than altering key semantics. 
Another observation related to Real-Guidance \cite{hesynthetic} is that it slightly alters images using SDEdit at very low strength. Although, it improves semantic alignment but it struggles with background. DA-Fusion \cite{trabucco2023effective}  also faces a similar  issue. Diff-Mix \cite{wang2024enhance} utilizes identifiers from other classes to transform the original image.
The method is prone to significantly modifying  the foreground  without effectively diversifying the background.
\par
Our method introduces more diversity while respecting local coherence and semantic alignment  with the training data. Our contextual prompts help the model to generate domain-specific image while negative prompts are responsible to restrict the T2I model from generating unrealistic image. Moreover, hard-cosine filtration sifts out low-confidence images in order to ensure that generated images are properly aligned.

\section{Training Batch}
\label{supp:batch_visulization}
To visually observe the diversity of training batch, we first visualize training batch of Mixup \cite{mixup}, CutMix \cite{cutmix}, SaliencyMix \cite{uddin2020saliencymix}, PuzzleMix \cite{kim2020puzzle}, GuidedMixup \cite{kang2023guidedmixup}, and \our{} approaches. 
\par
\begin{figure}
    \centering
    \includegraphics[width=1.0\linewidth]{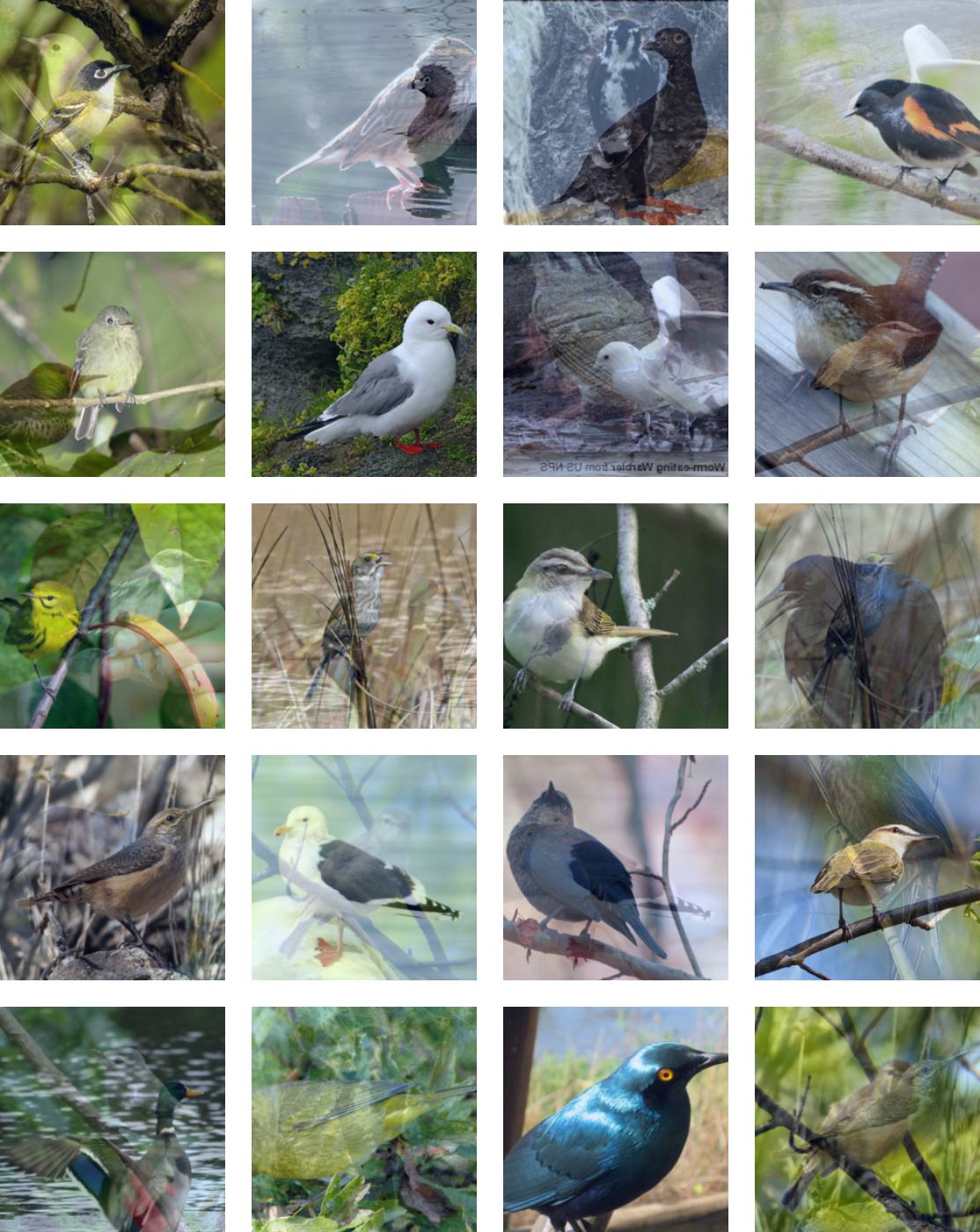}
    \caption{Batch visualization of Mixup \cite{mixup}}
    \label{fig:mixup_batch}
\end{figure}
\begin{figure}
    \centering
    \includegraphics[width=1.0\linewidth]{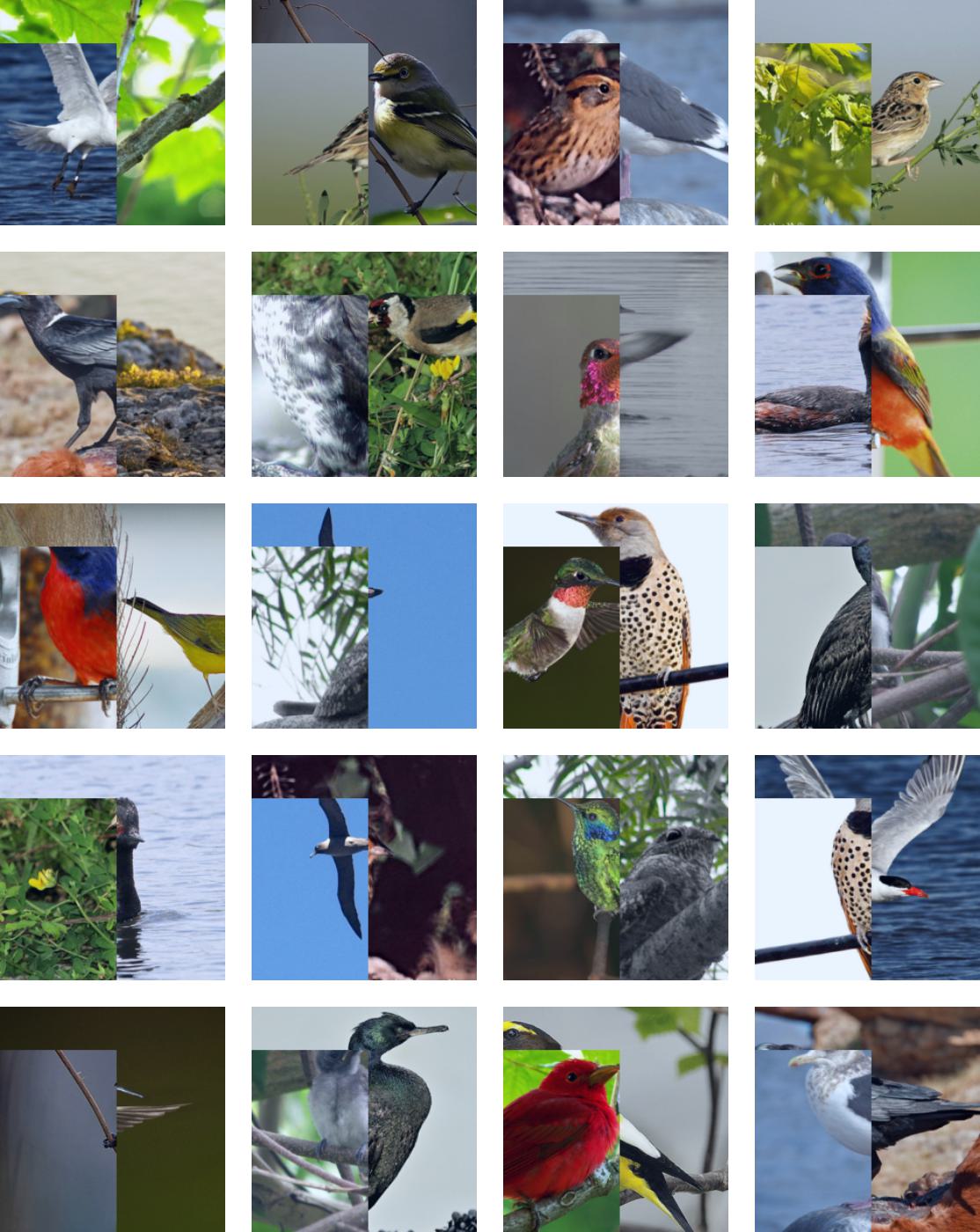}
    \caption{Batch visualization of CutMix \cite{cutmix}}
    \label{fig:cutmix_batch}
\end{figure}

In Figure \ref{fig:mixup_batch}, we observe the training batch of Mixup augmentation on CUB-Birds dataset, where source and target images are randomly blended using linear interpolation. Thus, introducing mixing source and target images with different $\lambda$ values. On the other hand, Figure \ref{fig:cutmix_batch} showcases the CutMix method, where random patches from the source image are placed on the target image to increase generalization and out-of-distribution detection. However, we observe primary limitation that introduces only patch-wise variations that may create unrealistic training data. Fortunately, the computational complexity of both Mixup \cite{mixup} and CutMix \cite{cutmix} is very low relative to  other baseline methods. 

\begin{figure}
    \centering
    \includegraphics[width=1.0\linewidth]{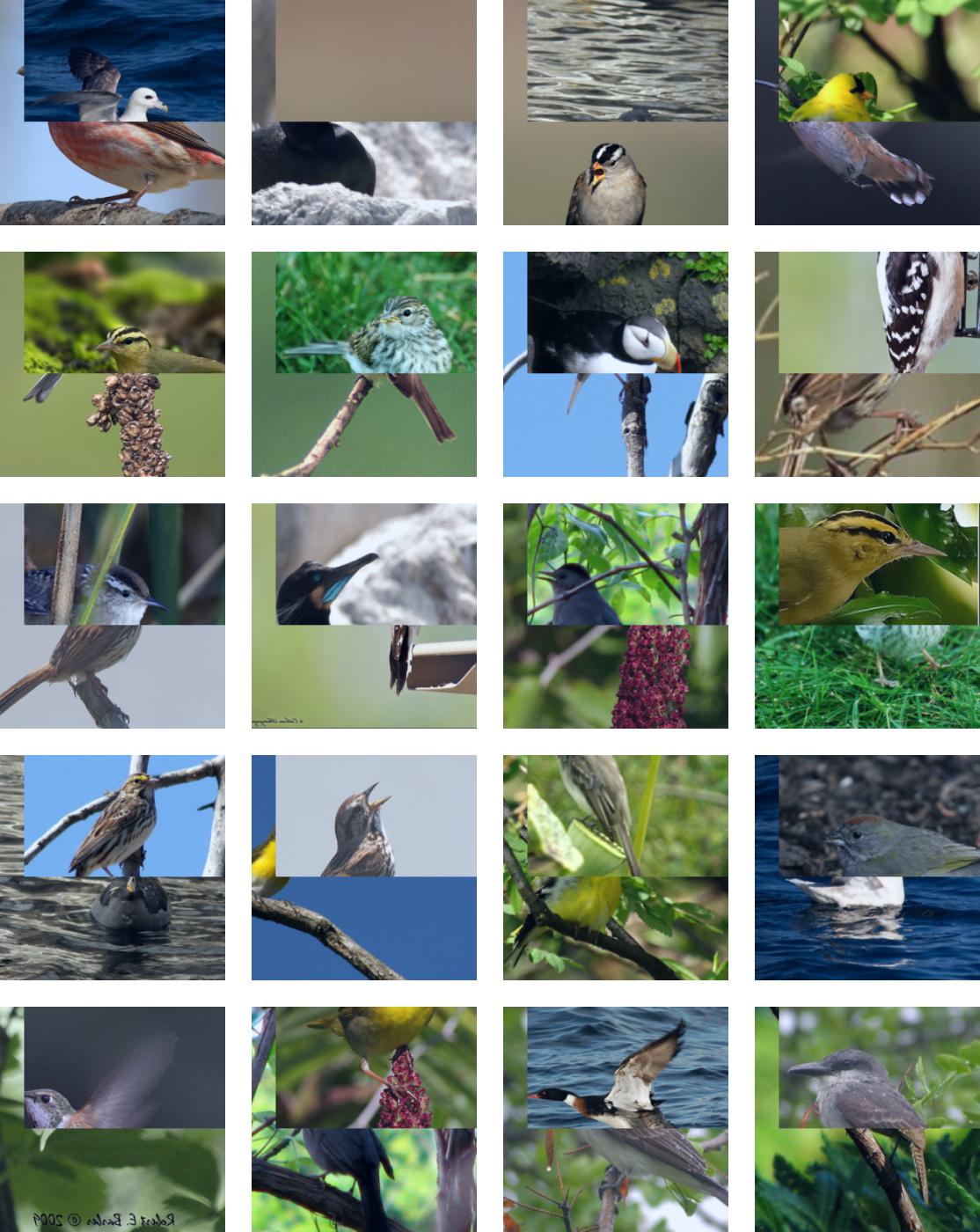}
    \caption{Batch visualization of SaliencyMix \cite{uddin2020saliencymix}}
    \label{fig:saliencymix_batch}
\end{figure}

\begin{figure}
    \centering
    \includegraphics[width=1.0\linewidth]{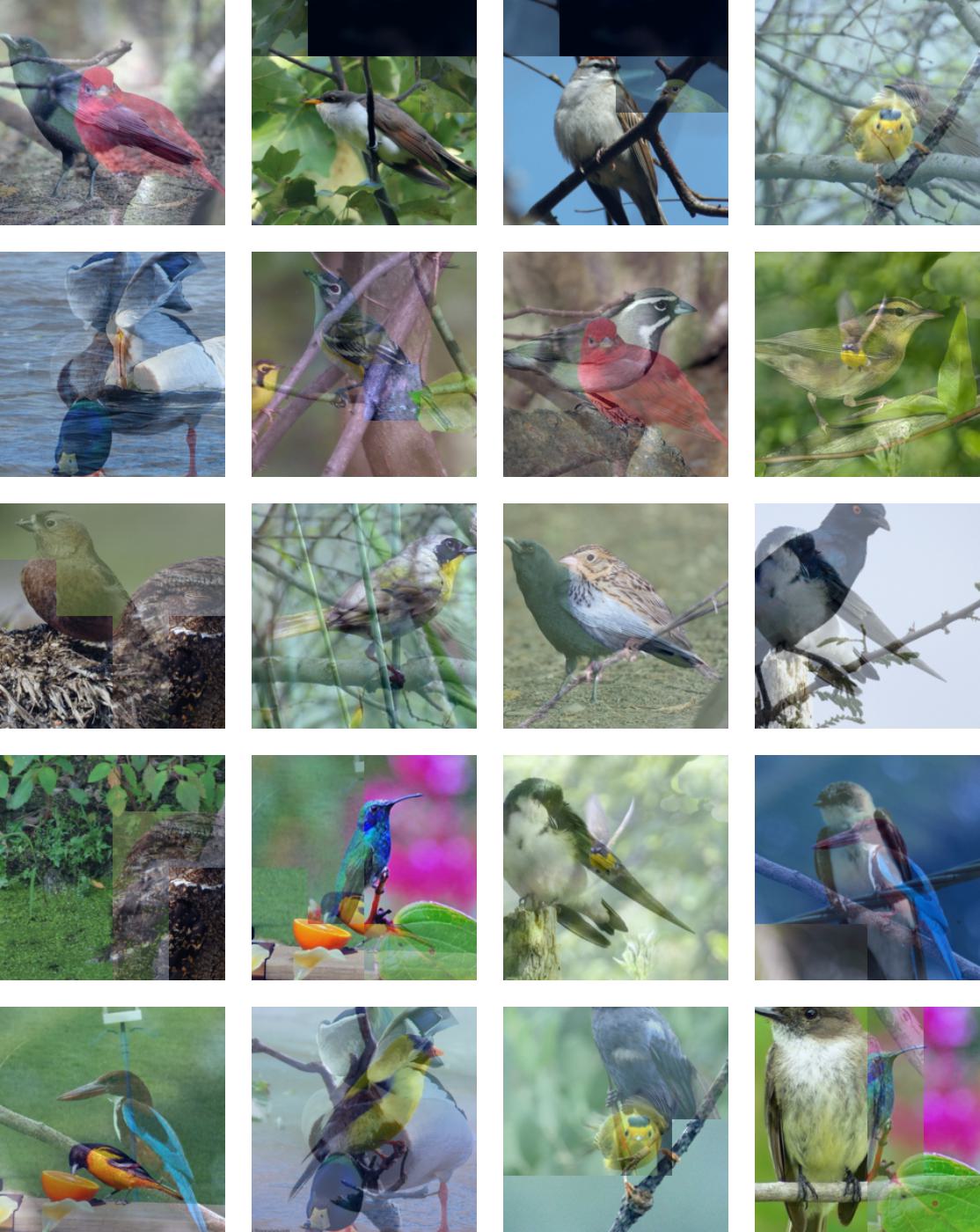}
    \caption{Batch visualization of PuzzleMix \cite{kim2020puzzle}}
    \label{fig:puzzlemix_batch}
\end{figure}

\begin{figure}
    \centering
    \includegraphics[width=1.0\linewidth]{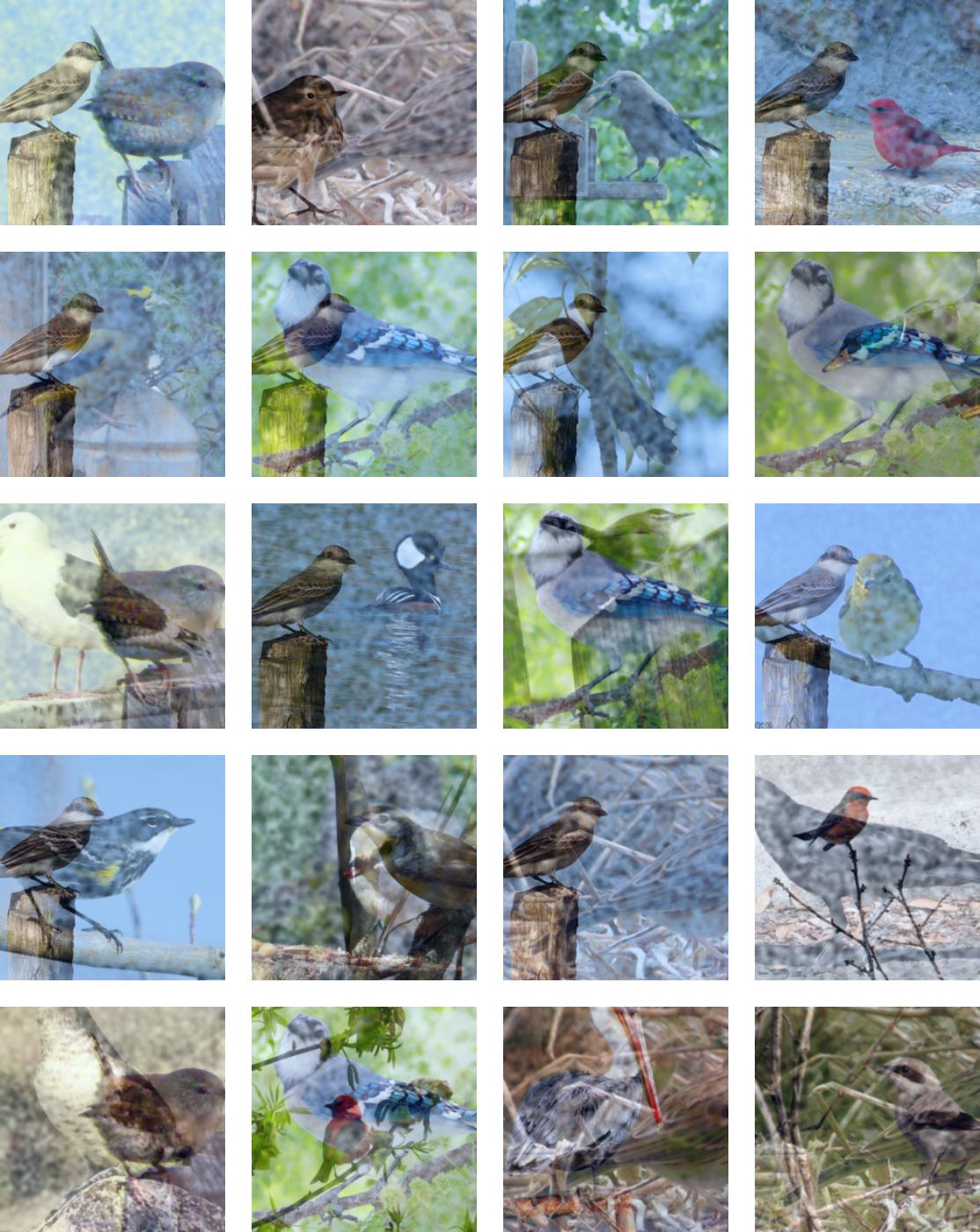}
    \caption{Batch visualization of GuidedMixup \cite{kang2023guidedmixup}}
    \label{fig:guidedap_batch}
\end{figure}

\begin{figure}
    \centering
    \includegraphics[width=1.0\linewidth]{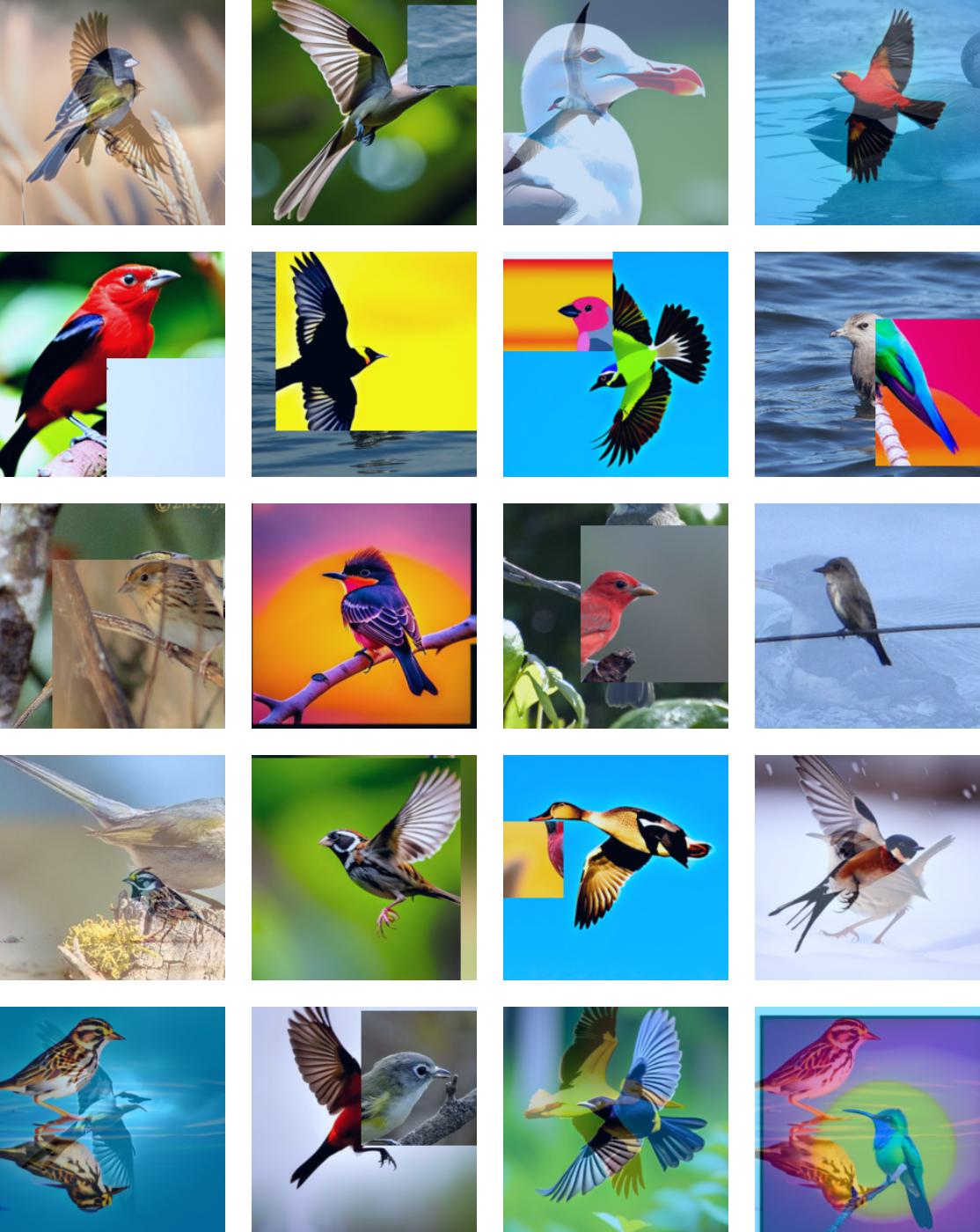}
    \caption{Batch visualization of our proposed method, \our{}}
    \label{fig:ours_batch}
\end{figure}

Shown in Figure \ref{fig:saliencymix_batch}, \ref{fig:puzzlemix_batch} and \ref{fig:guidedap_batch} are typical training  batches of saliency-based methods \cite{uddin2020saliencymix, kim2020puzzle, kang2023guidedmixup}. SaliencyMix \cite{uddin2020saliencymix}  identifies salient regions of the source images, and pastes it onto target image. In Figure \ref{fig:saliencymix_batch} batch, we identify that this method focuses on salient regions, such as areas with distinctive features or textures, and combines these regions with target images. However, it may not detect the salient region completely, as a result, it may miss crucial regions. Additionally, we also observe that it struggles with diversity since it primarily emphasizes on the most prominent features rather than a more varied combination of image regions. 
\par
Compare to SaliencyMix \cite{uddin2020saliencymix}, we notice that PuzzleMix \cite{kim2020puzzle} and GuidedMixup \cite{kang2023guidedmixup} optimize a spatial mixing strategy by selecting object patch  from various images and stitching them together, as  presented in Figure \ref{fig:puzzlemix_batch} and \ref{fig:guidedap_batch}. The methods leverage  both saliency and smoothness constraints to ensure the new augmented image retains contextually meaningful features from the original input data. Although, the results  showcase superior performance but also increase the computational overhead as the underlying approach involves optimizing patch selection based on various constraints.
\par
In our proposed method, \our{}, our batch contains more diverse samples as compared to previous methods \cite{trabucco2023effective, wang2024enhance} as seen in Figure \ref{fig:ours_batch}. 
Our objective is to preserve structural coherence while introducing new visual patterns for more robust model learning. In terms of balancing of realism and diversity, unlike previous methods \cite{trabucco2023effective, wang2024enhance}  that directly use real and generative image without any mixing strategies, \our{} maintains a balance by combining both aspects, which can help the model generalization.

\end{document}